\documentclass{article} %
\usepackage[preprint]{colm2026_conference}
\usepackage{minted}
\usepackage{listings}
\usepackage[dvipsnames]{xcolor}
\usepackage{algorithm}
\usepackage{algpseudocode}
\usepackage{subcaption}
\usepackage{amsmath}
\usepackage{makecell}
\usepackage{tcolorbox}
\usepackage{mdframed}
\usepackage{amssymb}
\tcbuselibrary{breakable, skins}
\newcounter{qacount}
\newtcolorbox{qabox}[2][]{
  breakable,
  enhanced,
  before upper = {\refstepcounter{qacount}},
  title        = {Example \theqacount\ (#2)},
  fonttitle   = \bfseries,
  colbacktitle= black!75,
  colframe    = black!75,
  colback     = blue!4,        %
  colbacklower= gray!8,        %
  segmentation style = {solid, black!40, line width=0.8pt},
  #1                           %
}
\floatname{listing}{Prompt}
\usepackage{graphicx}
\usepackage{microtype}
\usepackage{multirow}
\usepackage{hyperref}
\usepackage{cleveref}
\usepackage[title]{appendix} %
\usepackage{url}
\usepackage{booktabs}

\usepackage{lineno}

\definecolor{darkblue}{rgb}{0, 0, 0.5}
\definecolor{melon}{HTML}{F89E7B}
\hypersetup{colorlinks=true, citecolor=darkblue, linkcolor=darkblue, urlcolor=darkblue}

\title{Reasoning Models Know What's Important, \\and Encode It in Their Activations}

\author{
\textbf{Yaniv Nikankin}\textsuperscript{1},\quad
\textbf{Martin Tutek}\textsuperscript{2},\quad
\textbf{Tomer Ashuach}\textsuperscript{1},\\
\textbf{ Jonathan Rosenfeld}\textsuperscript{3},\quad
\textbf{Yonatan Belinkov}\textsuperscript{1,4}\\
\textsuperscript{1}Technion\quad
\textsuperscript{2}University of Zagreb, FER\quad
\textsuperscript{3}MIT\quad
\textsuperscript{4}Kempner Institute, Harvard\quad\\
}

\begin{document}

\ifcolmsubmission
\linenumbers
\fi

\maketitle
\vspace{-1em}
\begin{abstract}
Language models often solve complex tasks by generating long reasoning chains, consisting of many steps with varying importance. 
While some steps are crucial for generating the final answer, others are removable.
Determining which steps matter most, and why, remains an open question central to understanding how models process reasoning. 
We investigate if this question is best approached through model internals or through tokens of the reasoning chain itself.
We find that model activations contain more information than tokens for identifying important reasoning steps.
Crucially, by training probes on model activations to predict importance, 
we show that models encode an internal representation of step importance, even prior to the generation of subsequent steps.
The internal representations of importance in different models yield high agreement on which steps are important.
The representation is distributed across layers, and does not correlate with surface-level features, such as a step's relative position or its length.
Our findings suggest that analyzing activations can reveal aspects of reasoning that surface-level approaches fundamentally miss, indicating that reasoning analyses should look into model internals.\footnote{Code and data available at \url{https://github.com/technion-cs-nlp/reasoning-importance}}

\end{abstract}

\vspace{-0.5em}
\section{Introduction}
\label{sec:intro}

Reasoning language models solve complex problems by generating long reasoning chains that consist of many intermediate steps \citep{jaech2024openai, guo2025deepseek, agarwal2025gpt, team2025kimi}. 
Some of these steps are critical for the final answer, while others are redundant or uninformative, making them removable \citep{qu2025survey}.
Identifying which steps are truly \emph{important} has significant implications for reasoning efficiency, faithfulness, and interpretability: 
it determines whether the model's answer is grounded in its reasoning chain, and which steps actually drive the answer.

Identifying important steps can be done by inspecting the tokens of the reasoning chain or by analyzing internal activations.
Previous studies have indirectly examined step importance in both ways \citep{bogdan2025thoughtanchors, chen2026decoupling, levy2025humans, do2025definesgoodreasoningllms, zhang2025reasoning}.
These approaches are complementary: while token-based approaches are often simpler, they are not always faithful \citep{lanham2023measuring} and may lose valuable information in the transition between latent activations and sampled tokens.
In this work, we study reasoning step importance using both approaches and compare their utility.

\begin{figure}[t]
    \centering
    \includegraphics[width=.95\linewidth]{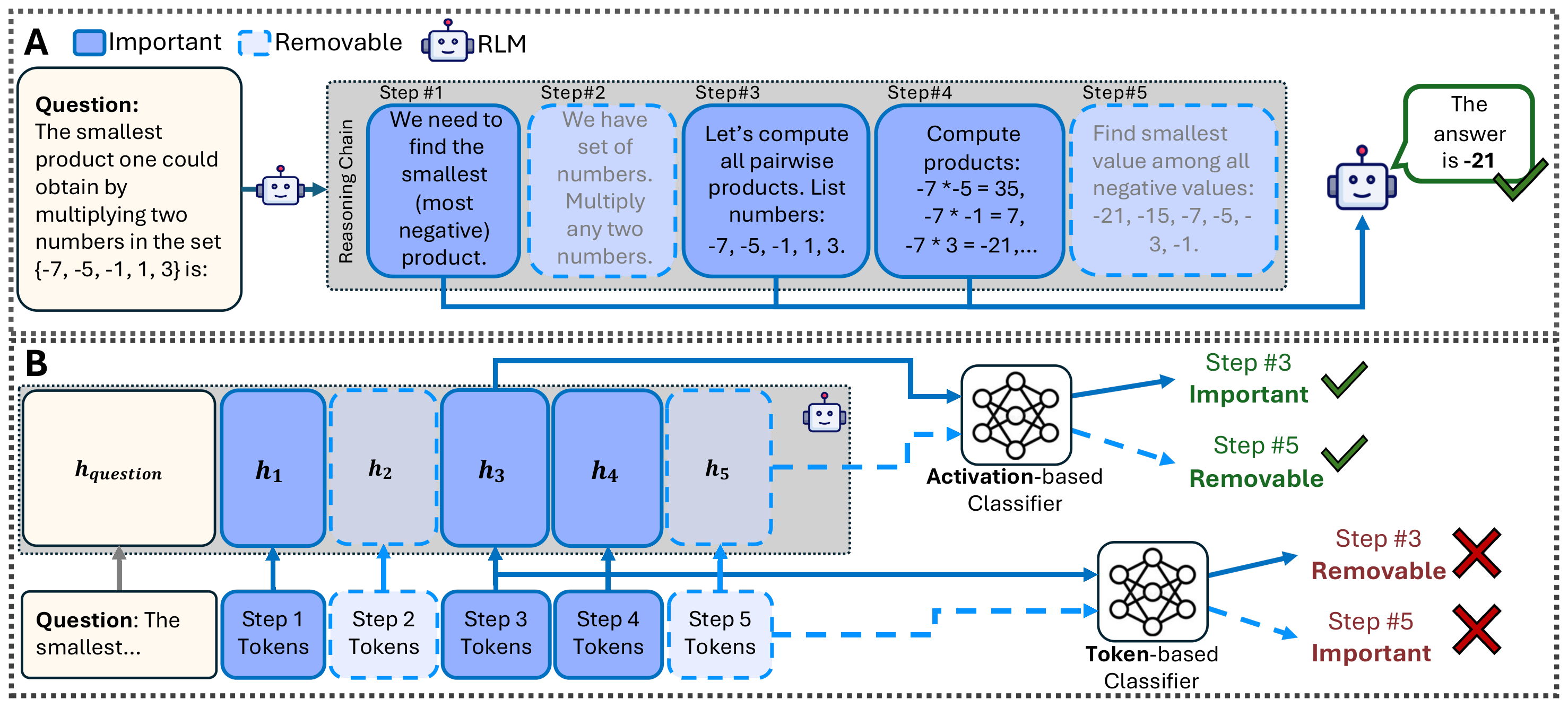}
    \caption{\textbf{Overview of reasoning step importance.}
    \textbf{A.} A reasoning chain can be partitioned into important and removable steps. Important steps make up a sparse subsequence that is sufficient to reach the correct answer. Removable steps can be omitted.
    \textbf{B.} The importance of a reasoning step is encoded in the model's latent activations, and can be extracted using a probe, even before subsequent steps are generated.
    In contrast, this importance signal can not be accurately distinguished from the reasoning chain alone.
    }
    \label{fig:opening-figure}
\end{figure}

To see if model activations are more indicative of importance compared to tokens, we analyze mathematical reasoning across five models.
Following \citet{tutek2025measuring}, we define importance via \textit{removability}: a step is important if removing it prevents the model from reaching the correct answer.
For each reasoning chain, we greedily remove steps ranked by either latent influence signals (activation-based) or an external LLM judge (token-based).
We find that the activation-based approach consistently identifies a smaller set of steps that is sufficient for generating the correct answer (\Cref{fig:opening-figure}A), suggesting that latent information leads to more precise identification of reasoning importance.

More interestingly, we ask whether models encode an \emph{internal} notion of step importance.
By training probes on latent activations of individual steps, we find that important steps are internally distinguishable from removable ones (\Cref{fig:opening-figure}B), suggesting models represent importance in their activations.
This representation relies on activations that integrate the full reasoning history rather than local step content alone.
Remarkably, this representation exists even \emph{prior} to the generation of subsequent steps, indicating that reasoning models encode importance pre-hoc.
This internal representation yields high agreement across models: the internals of one model can be used to predict importance labels from another model.
This internal representation exhibits low correlation with surface-level features, such as position, length and step category, suggesting importance is not surface-level.

Together, our results show that LLMs encode a notion of reasoning step importance within a step's activations---even prior to the generation of subsequent steps---a notion that does not surface in token outputs.
This suggests that to better understand different aspects of reasoning in LLMs, one must look beyond tokens and into model internals.

\section{Preliminaries}
\label{sec:prelims}
We study the concept of reasoning step importance, and how it is encoded in reasoning models, through the proxy of removability.
We consider reasoning steps important if they can not be removed while maintaining the correct answer (see discussion in \Cref{app:removability-discussion}).

\vspace{-0.5em}
\subsection{Problem setup}
\label{sec:problem-setup}

For a prompt $x$ and a model $\phi$, we denote $o = \phi(x) = s_1 \circ s_2 \circ\cdots\circ s_n$ as its corresponding reasoning chain of $n$ steps, where $\circ$ denotes concatenation.
A reasoning chain subsequence $\mathcal{T} = \{s_{i_k}\} = (s_{i_1}, s_{i_2}, \ldots)$, where $i_k < i_{k+1}$ for all $k$, is any subsequence of steps of $o$.
We evaluate the quality of reasoning subsequences using \emph{sufficiency} and \emph{attainability}.

\textbf{Sufficiency} measures how well a reasoning subsequence supports generation of a correct final answer.
We extend a subsequence $\mathcal{T}$ with an answer-elicitation string $\sigma$ (e.g., ``\texttt{So the answer is \textbackslash boxed\{}''; see \Cref{app:technical-details-generation-segmentation-evaluation}) and sample $K=5$ non-deterministic completions.
Sufficiency is the fraction of completions that yield the ground truth answer $a^*$:
\begin{equation}
    \mathrm{Sufficiency}(\mathcal{T}) = \frac{1}{K} \sum_{k=1}^{K} \mathbf{1}\!\left[\phi(x \circ \mathcal{T} \circ \sigma) = a^*\right]\ .
\end{equation}

\textbf{Attainability} measures how naturally a reasoning subsequence arises from the model's prior---i.e., how probable it is given the input.
We measure this via the perplexity of $\mathcal{T}$ conditioned on $x$, normalized to $[0,1]$ and reversed, so that higher scores indicate more probable subsequences.
We normalize using two reference bounds. 
The lower bound is the perplexity of the full chain $o$, a highly probable continuation of the prompt $x$.
The upper bound is the perplexity of the direct-answer string $\sigma \circ a^*$: the model is ``surprised'' to see the correct answer immediately after a hard problem statement.
Formally:
\begin{equation}
    \mathrm{Attainability}(\mathcal{T}) = 1 - \frac{\mathrm{PPL}(\mathcal{T} \mid x) - \mathrm{PPL}(o \mid x)}{\mathrm{PPL}(\sigma \circ a^* \mid x) - \mathrm{PPL}(o \mid x)},
\end{equation}
where $\mathrm{PPL}(\cdot \mid x)$ is the perplexity assigned by the model to a sequence given context $x$.

\vspace{-0.2em}
The two metrics are complementary by design. 
Sufficiency alone is susceptible to degenerate subsequences: a subsequence consisting solely of the sentence ``$\texttt{The answer is } a^*$'' would be sufficient, since the model can simply copy the answer---yet such a subsequence is unattainable for a hard problem.

For each reasoning chain, we search for the important steps that make up a \textbf{core reasoning subsequence}: a subsequence $\mathcal{T}^*$ that is sufficient, attainable, and as short as possible.
Formally, $\mathcal{T}^*$ is the shortest valid subsequence from which no single step can be removed while preserving sufficiency and attainability:
\begin{equation}
     \mathcal{T}^* = \arg\min_{\mathcal{T} \subseteq o} |\mathcal{T}| \quad \text{subject to} \quad \mathrm{Sufficiency}(\mathcal{T}) \geq \tau \ \text{ and } \ \mathrm{Attainability}(\mathcal{T}) \geq \tau \ ,
\end{equation}
where $\tau=0.8$, corresponding to $\frac{4}{5}$ correct answers and low relative perplexity (see App~\ref{app:suff-atnb-threshold-ablation}).

\subsection{Experimental details and pre-processing}
\label{sec:experimental-details}

\paragraph{Models and datasets.}
We perform our analysis on five reasoning models from several families, differing in pretraining data, architecture, and size: GPT-OSS-20B \citep{agarwal2025gpt}, DeepSeek-R1-Distill-Qwen (1.5B, 7B, and 14B) \citep{guo2025deepseek}, and OLMo-3-7B-Think \citep{olmo2025olmo}.
We examine the Human Annotated Reasoning Problems \citep[HARP;][]{yue2024harp} and MATH-500 \citep{hendrycks2021measuring} datasets, two challenging mathematical reasoning benchmarks containing $4,780$ and $500$ problems, respectively.

\paragraph{Generation.}
For each prompt $x$, we generate a reasoning chain $o$ using greedy decoding.
Following prior work \citep{lanham2023measuring, jacovi2024chain, bogdan2025thoughtanchors}, we segment $o$ into steps analogous to sentences.
We validate each chain by comparing the final answer $\phi\left(x\right)$ to the ground truth $a^*$ using symbol-matching heuristics, since correct mathematical answers may be expressed in multiple equivalent forms.
We provide full details in \Cref{app:technical-details-generation-segmentation-evaluation}, with example reasoning chains shown in \Cref{app:reasoning-chain-examples}.

\paragraph{Filtering.}
We restrict our analysis to prompts for which the reasoning chain is both correct and necessary.
To measure necessity, we check whether the model can produce a correct answer without reasoning by prompting it with $x \circ \sigma$. 
We draw $K=5$ non-deterministic completions and compute the percentage of correct answers.
A reasoning chain is considered necessary for a prompt $x$ if the model answers incorrectly at least $4$ out of $5$ resamples.
This ensures that every retained prompt genuinely requires multi-step reasoning.
We report the number of filtered prompts in \Cref{app:performance-stats}.

To reduce the search space of potential important steps in each reasoning chain and account for redundant reasoning paths, we employ two additional pre-processing steps per prompt: we remove shortcut steps and identify a reasoning prefix sufficient to reach the final answer.

\paragraph{Eliminating shortcut steps.} \vspace{-0.2em}
A step $s_i$ is considered a \emph{shortcut step} if it alone is sufficient for producing the correct answer, either because $\mathrm{Sufficiency}(s_i)$ is high or because it states the answer directly (e.g., ``\texttt{So the answer is $-21$}'', see App~\ref{app:technical-details-generation-segmentation-evaluation}).
Including such steps conflates answer-stating with meaningful reasoning, so we remove them from each chain.
We then verify that the remaining chain is still sufficient, and exclude chains that fail this check.

\paragraph{Identifying a sufficient prefix.} 
Reasoning models often reach an answer long before outputting it in the reasoning chain \citep{mao2025early, wang2025chain}.
Thus, rather than analyzing the full chain, we restrict our analysis to a \emph{sufficient prefix}: the shortest contiguous subsequence $\mathcal{T}_p = (s_1, s_2, \ldots, s_p)$ that achieves both high sufficiency and attainability. 
This serves two purposes.
First, it discards trivially removable post-answer content (e.g., verification).
Second, it ensures that we analyze a single coherent path from the input $x$ to the answer, rather than a chain that may contain multiple redundant sufficient subsequences.

\section{Finding important reasoning steps}
\label{sec:method}
Having identified a reasoning chain prefix $\mathcal{T}_p$, we now aim to find a core reasoning subsequence $\mathcal{T}^*\subseteq \mathcal{T}_p$.
Since exhaustive search over all $2^p$ subsequences is infeasible, we use a greedy procedure instead.
We consider three variants for this procedure: an activation-based variant, a token-based variant, and a randomized variant.

\subsection{Methods}
The greedy procedure operates in two steps.
First, we select steps from $\mathcal{T}_p$ that have direct or indirect influence on the answer, forming $\mathcal{T}_{\mathrm{inf}}$.
Second, we rank steps in $\mathcal{T}_{\mathrm{inf}}$ and iteratively remove them, while maintaining sufficiency and attainability of the resulting subsequence, yielding the core subsequence $\mathcal{T}^*$.
The three variants differ in how they measure influence in the first stage and rank steps in the second stage.

\paragraph{Activation-based variant.}
To identify the steps that influence the answer, we calculate the influence between every pair of steps $s_i,s_j$, $j>i$. 
We do so with gradient attribution \citep[][;see \Cref{app:attribution-comparison}]{li2016visualizing}.
We sum the probabilities of all tokens in $s_j$ and compute the gradient of this sum with respect to the input embeddings of all tokens in $s_i$.
The influence score is the Frobenius norm of the resulting gradient tensor:
\begin{equation}
    \mathbf{I}(s_i, s_j) = \left\| \frac{\partial \sum_{t \in s_j} \mathbb{P}_{t}}{\partial \mathbf{E}_{s_i}} \right\|_F,
\end{equation}
where $\mathbf{E}_{s_i}$ is the matrix of input embeddings for step $s_i$ and $\mathbb{P}_t$ is the probability of token $t$. 
We set the influence of a step on itself to zero.
This yields an influence matrix $\mathbf{I}$ over all pairs of steps $(s_i, s_j)$ with $j>i$, which we compute in a single backward pass per target step.

In the first stage, we use $\mathbf{I}$ to identify steps that have direct or indirect influence on the answer. 
We initialize $\mathcal{T}_{\mathrm{inf}} = (s_p)$ ($s_p$ is the required final prefix step; see \Cref{sec:prelims}).
For each $s_j \in \mathcal{T}_{\mathrm{inf}}$, we recursively add all predecessor steps $s_i$ whose cumulative influence $\sum_{s_i}\mathbf{I}(s_i, s_j)$ exceeds a threshold. 
We sweep over thresholds $\{0.1, 0.2, \ldots, 1.0\}$, selecting the lowest one such that $\mathcal{T}_{\mathrm{inf}}$ maintains high sufficiency and attainability.

In the second stage, we rank the remaining steps in $\mathcal{T}_{\mathrm{inf}}$ by their mean influence on subsequent steps.
We attempt removals from lowest to highest rank, and retain a removal only when the resulting subsequence still satisfies the sufficiency and attainability thresholds.
The full implementation of the activation-based approach is presented in \Cref{app:activation-based-labeling-procedure}.

\paragraph{Token-based variant.}
Rather than using internal model signals, the token-based method relies entirely on an external LLM (Gemini-2.5-Pro \citep{comanici2025gemini}). %
In the first stage, we prompt the LLM to identify which steps in $\mathcal{T}_p$ influence the answer, yielding $\mathcal{T}_{\mathrm{inf}}$.
In the second stage, we prompt the LLM to rank the steps in $\mathcal{T}_{\mathrm{inf}}$ by importance and use this ranking in place of the mean influence score.
The prompts used in this variant are presented in \Cref{app:token-based-labeling-procedure}.

\paragraph{Random variant.}
We randomly rank steps for removal.
This variant serves as a strong baseline: any improvement of the other variants over it can be attributed to the quality of their respective rankings.
We skip the first stage, noting $\mathcal{T}_{\mathrm{inf}}=\mathcal{T}_p$.
In the second stage, steps in $\mathcal{T}_{\mathrm{inf}}$ are ranked in a random order, rather than by a meaningful signal.\footnote{Because all steps are removal candidates, this variant requires significantly more runtime.} %

\begin{table}[t]
\centering
\begin{tabular}{llrrrrr}
\toprule
Dataset & Method & GPT-OSS & DS-1.5 & DS-7 & DS-14 & Olmo3 \\
\midrule
\multirow{3}{*}{HARP} & Random Greedy & 26.2 & 29.7 & 29.2 & 32.1 & \underline{24.1} \\
 & Token-based & 31.7 & 41.0 & 41.4 & 40.7 & 35.3 \\
 & Activation-based & \underline{23.0} & \underline{26.4} & \underline{27.0} & \underline{28.1} & 26.3 \\
\midrule
\multirow{3}{*}{Math-500} & Random Greedy & 33.4 & 26.0 & 24.8 & 30.0 & 21.8 \\
 & Token-based & 42.5 & 31.4 & 35.0 & 41.6 & 34.1 \\
 & Activation-based & \underline{30.8} & \underline{22.1} & \underline{23.1} & \underline{27.8} & \underline{18.0} \\
\bottomrule
\end{tabular}
\caption{\textbf{Average core reasoning subsequence size (\%, \underline{lower} is better).} 
    We measure the average size of the core reasoning subsequence $\mathcal{T}^*$, as a percentage of the sufficient reasoning prefix $\mathcal{T}_p$.
    Across models and datasets, the activation-based method produces smaller subsequences (underlined), suggesting it better identifies the steps that matter most.}
\label{tab:pruning-comparison}
\end{table}

\subsection{Results} 
We evaluate each variant by the size of the core reasoning subsequence $\mathcal{T}^*$ it produces.
Results in \Cref{tab:pruning-comparison} show that the activation-based variant consistently yields the smallest core subsequences.
Detailed per-prompt comparisons in \Cref{app:more-per-prompt-variant-comparison} also point to a consistent advantage of the activation-based variant.%
We therefore adopt the activation-based approach for all subsequent analyses, using it to assign binary labels to steps in $\mathcal{T}_p$: steps removed in the first stage are labeled \emph{removable}, and steps retained after the second stage are labeled important and \emph{non-removable}.
Steps removed during the second stage are left unlabeled; see \Cref{app:removability-discussion} for a discussion on this choice.

\section{Detecting the notion of importance}
\label{sec:identifying-importance}

Having found core reasoning subsequences, we now assess if we can distinguish important from removable reasoning steps using their representations.
Specifically, we ask if models maintain an internal notion of importance, and if so, what kind of representation, from tokens to context-aware activations, is sufficient to detect it?

\subsection{Models encode importance in their activations}
\label{sec:activation-importance-exp}

\begin{figure}[t]
    \centering
    \includegraphics[width=\linewidth]{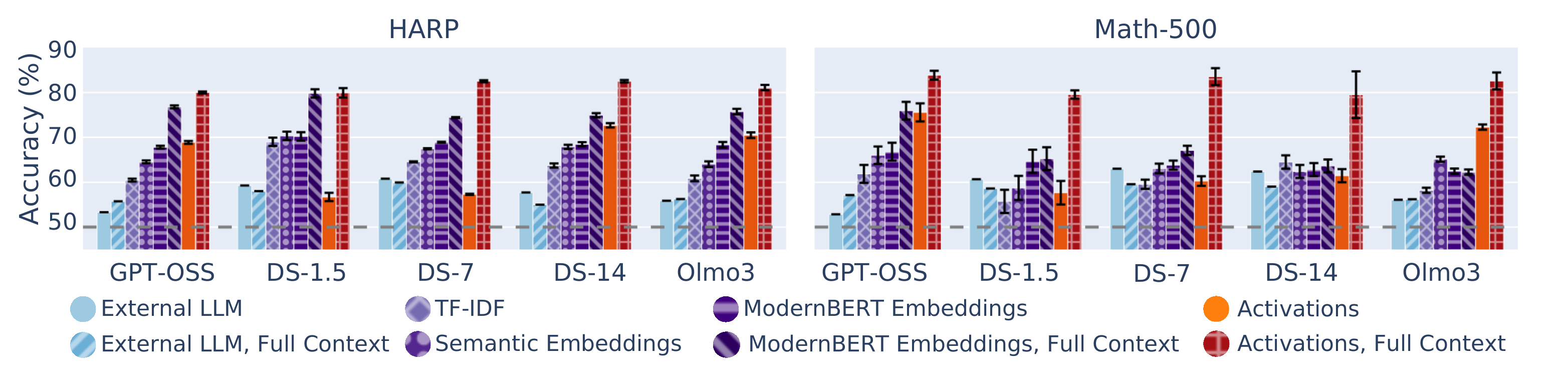}
    \caption{\textbf{Importance detection requires full context and strong representations.}
    We compare methods for predicting reasoning step importance pre-hoc, ranging from token-based approaches to full-context activations.
    Token-based LLM-as-a-judge, with or without full context, reaches low accuracies ($\leq 60\%$).
    Methods using context-free or weak representations (TF-IDF, semantic or ModernBERT embeddings, and context-free activations) perform moderately.
    Full-context activation probes achieve the highest accuracies ($\geq 80\%$), confirming that both full context and rich representations are necessary to accurately detect reasoning step importance.
    }
    \label{fig:importance-classification}
    \vspace{-1em}
\end{figure}

We begin by analyzing if models maintain an internal notion of importance in their activations.
To test this, we train a binary classification probe on the activations of each reasoning step to predict if it is removable, using labels derived from the best-performing activation-based variant in \Cref{sec:method}.
We pass full prompts, including the input problem and the sufficient prefix, through the model. 
We extract activations at the positions of each target reasoning step.
The activations of a step form a tensor of shape $S \times L \times d$, where $S$, $L$, and $d$ are the number of tokens, layers, and the hidden dimension, respectively.
We average the activation tensor over the token dimension and flatten it over the layers and hidden dimensions.
As the probe, we use a two-layer MLP with a hidden dimension of $128$, and a ReLU non-linearity; alternative architectures are evaluated in App~\ref{app:importance-classifier-technical-details}.
Probes are trained and evaluated on a balanced set of removable and non-removable steps on $3$ random splits.

We find that activation-based probes (red bars in \Cref{fig:importance-classification}) yield high ($\geq 80\%$) accuracies, indicating that models internally encode a signal for reasoning step importance
\footnote{As a control against circularity, we repeat this with other labels in \Cref{app:probe-performance-random-labeling}}
.
Notably, since a step's activations contain information only from that step and preceding steps (due to causal attention), this notion of importance is \emph{pre-hoc}: it exists before the model generates the subsequent steps that rely on the current step and thereby determine its importance. 

\subsection{Detecting importance requires a strong representation and context}
\label{sec:baselines-for-importance-detection}
Having established that full-context activation probes successfully identify importance pre-hoc, we ask whether simpler extraction methods can suffice.
The baselines we consider differ along two dimensions: context availability and the representational capacity---the richness of each reasoning step's encoding. 

On the context side, we evaluate a context-free variant of the activation probe, in which only the target step is encoded by the model without any preceding reasoning.
To measure if importance can be detected from weaker step representations, we train classifiers to predict step importance from three step representations with different capacities:
(i) TF-IDF representation over n-grams ($n \in [1,5]$) \citep{SprckJones2021ASI}, capturing surface lexical patterns; 
(ii) semantic embeddings from a pre-trained sentence transformer \citep[\texttt{all-MiniLM-L6-v2};][]{reimers2019sentence}, capturing sentence-level semantics; and
(iii) output embeddings from ModernBERT \citep{warner2025smarter} finetuned for importance prediction, with or without context.
In all cases, the classifier's architecture is identical to the activation probe, so any accuracy gap reflects representation quality rather than classifier capacity.
Finally, we evaluate a strong external LLM (Gemini-2.5-Pro) as a zero-shot judge in both context-free and full-context settings.
We prompt it to classify each reasoning step\footnote{Due to budget constraints, LLM-as-a-judge evaluations use $1500$ steps per model and dataset. Results of other methods on this subset are consistent with full comparisons (\Cref{app:more-results-for-classification}).}
as removable or non-removable, using classification prompts shown in \Cref{app:baselines-for-importance-classification}.
This method operates purely over tokens as a representation of reasoning steps.

The results (\Cref{fig:importance-classification}) are consistent across models and datasets.
Removing context from the activation probe causes a substantial accuracy drop ($10$--$20\%$, orange bar), establishing that the input prompt and the preceding reasoning chain are necessary to identify step importance. This suggests that methods that encode reasoning steps in isolation \citep{bogdan2025thoughtanchors, levy2025humans} might miss such signals.
Reducing representational capacity also decreases performance: the TF-IDF, semantic embeddings and ModernBERT approaches all trail the full-context activation probe (purple bars).
The context-aware ModernBERT scores higher than its context-free counterpart, validating that access to full context is critical for identifying step importance.
Finally, the LLM-as-a-judge reaches low accuracies ($\leq 60\%$, blue bars), even with full context.
This indicates that token-level access alone is likely insufficient, as even a strong reasoning model does not perform the task well without access to latent representations.

Taken together, these results confirm that accurately detecting reasoning step importance requires access to both strong latent representations and full context.
Both requirements are satisfied in the activations of the model that generated the reasoning chain, further motivating the use of full-context activations for investigating reasoning.

\section{Analyzing the notion of importance}
\label{sec:understanding-importance}
Having established that importance is encoded in full-context model activations, we now analyze the nature of this encoding: how does it vary across models, where is it encoded in activations, and which other features does it correlate with.

\subsection{Importance Is cross-model}
\label{sec:cross-model-importance}
\begin{figure}[t]
    \centering
    \includegraphics[width=\linewidth]{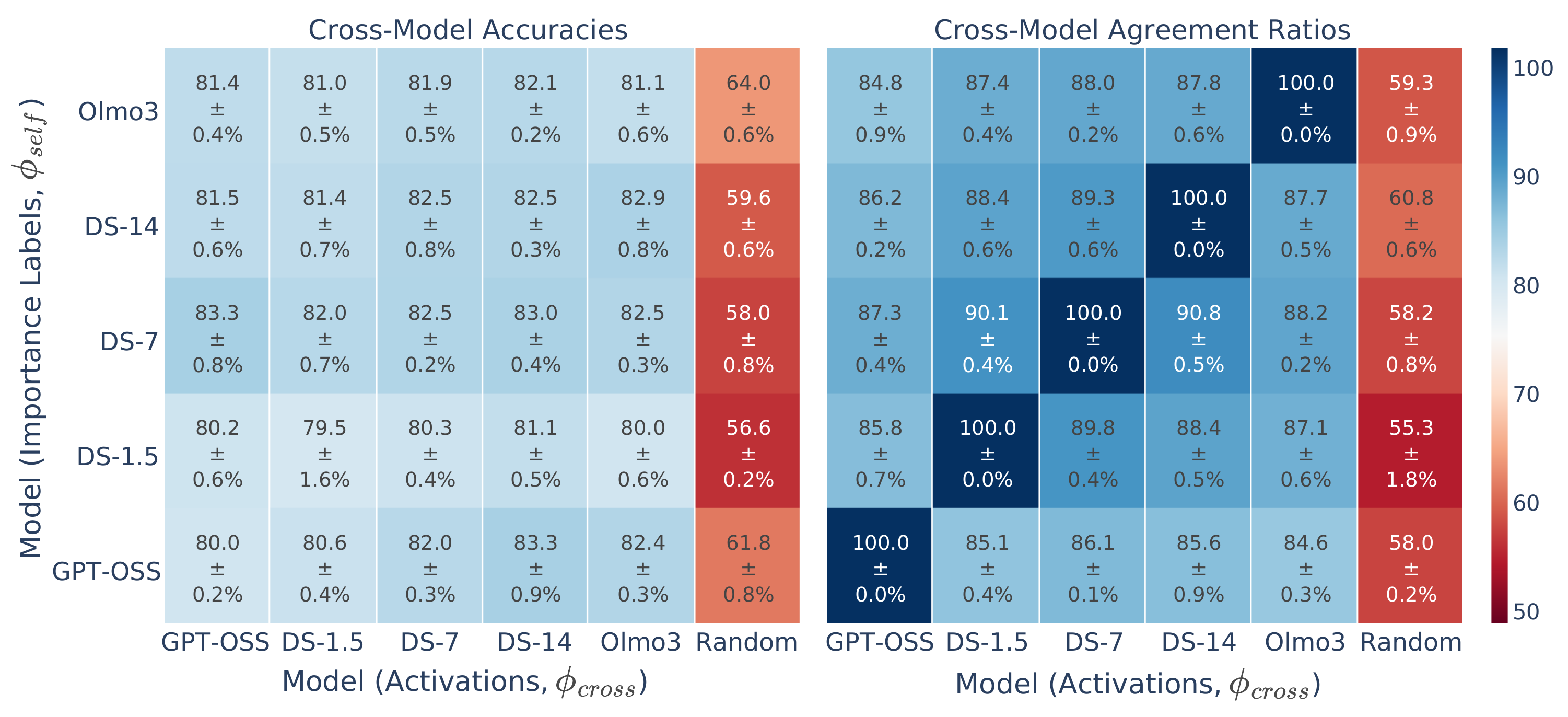}
    \caption{\textbf{Activation-based probes (on HARP) perform similarly across models.}
    \textbf{Left:} Probes trained on activations from one model ($\phi_\text{cross}$, x axis) to predict importance labels from another ($\phi_\text{self}$, y axis) achieve high accuracy across all pairs.
    \textbf{Right:} 
    We train two activation probes, on activations from $\phi_\text{cross}$ and $\phi_\text{self}$, both with labels from $\phi_\text{self}$. 
    The fraction of reasoning steps assigned the same label by both probes is high across all pairs, suggesting importance is encoded consistently across models.
    The \textit{random} columns report results of a randomly initialized model matching $\phi_\text{self}$ to verify that probe performance depends mostly on informative activation features rather than spurious correlations.
    }
    \label{fig:cross-model-results}
\end{figure}

Prior work has shown that reasoning chains can generalize across models \citep{pal2026explanations}, yet others posited that models explain their own computations better than other models can \citep{li2025training}.
We ask if the internal notion of importance generalizes across models: can a probe trained on activations from one model predict importance labels from another?

For every pair of models $\phi_\text{self}$, $\phi_\text{cross}$, we pass reasoning chains generated by $\phi_\text{self}$ through $\phi_\text{cross}$ to obtain activation tensors.
We train a probe on activations from $\phi_\text{cross}$ to predict importance labels of $\phi_\text{self}$, using a two-layer MLP probe trained on full context activations. 
We train and evaluate over $3$ random splits across all labeled reasoning steps from $\phi_\text{self}$. 
As a baseline, we replace $\phi_\text{cross}$ with a randomly initialized model matching the architecture of $\phi_\text{self}$ \citep{tenney2019you}, and train a probe on its activations following the same procedure.

For each model pair, we measure cross-model probe accuracy and agreement ratio.
The agreement ratio is the percentage of steps classified identically by the probe trained on $\phi_\text{self}$ activations and the probe trained on $\phi_\text{cross}$ activations.
We report results for HARP (\Cref{fig:cross-model-results}) and MATH-500 (\Cref{app:math-crossmodel-additional-results}).
Across all model pairs, the cross-model probe accuracy matches the within-model setting, and agreement ratios are high.
In contrast, the randomized baseline reaches lower but non-random accuracy, confirming that probe performance \emph{mainly} relies on meaningful activation structure, which does not appear in randomized weights.
These results indicate that the importance of a reasoning step is consistently encoded in activations regardless of which reasoning model processes it, suggesting importance is not model-specific but a more universal concept.

\subsection{Importance is represented by distributed, non-surface-level features}
\label{sec:concept-analysis}

\begin{figure}[t]
    \centering
    \includegraphics[width=\linewidth]{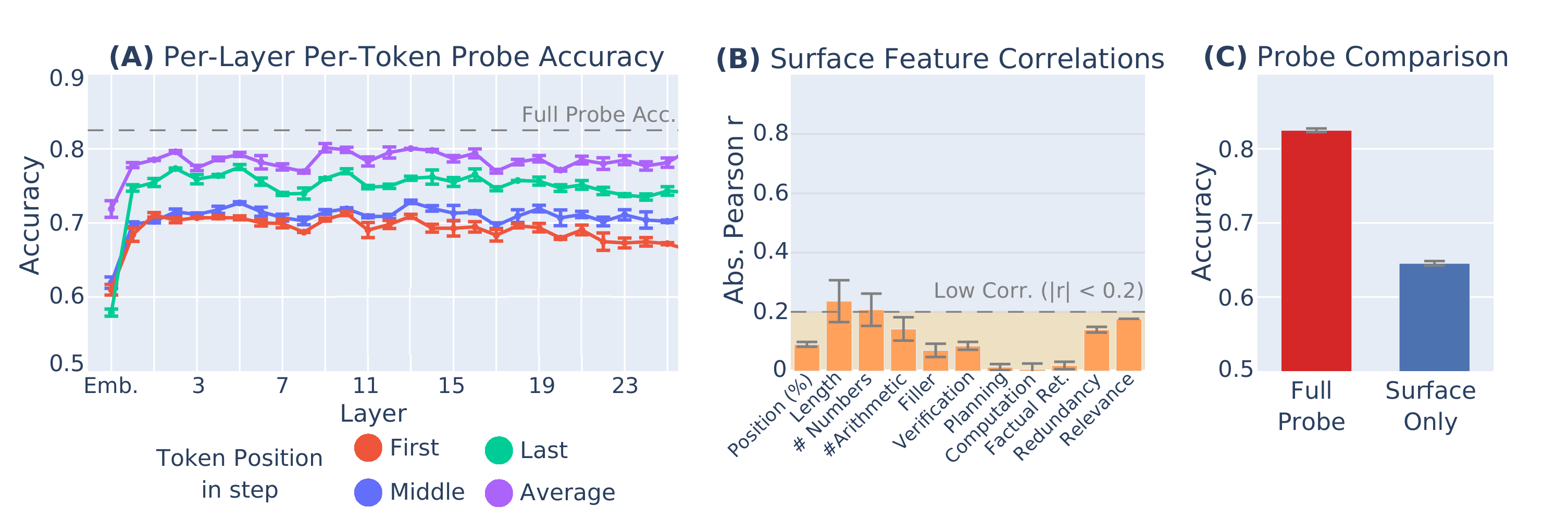}
    \caption{\textbf{Analysis of the importance probe in DeepSeek-7B (HARP dataset).}
    \textbf{(A)} We train separate probes on residual activations at each layer and token position; no single layer or position matches the accuracy of the full probe trained on all layers. 
    \textbf{(B)} 
    Individual surface-level features show low correlation with probe predictions;  
    \textbf{(C)}
    a regression model trained on all surface features (combined) performs substantially worse than the full probe, indicating that the probe captures meaningful structure beyond surface-level information.
    }
    \label{fig:deepseek-7b-probe-analysis}
    \vspace{-0.5em}
\end{figure}

We analyze the importance probe by examining two questions: is importance localized to specific layers and positions, and whether it reduces to surface-level properties of the text.

\paragraph{Importance is distributed across layers and positions.}
To isolate where importance is encoded, we train a separate probe on the residual-stream activations of each individual layer, at specific token positions of each step (first, middle, and last).
We also train a per-layer probe over activations averaged across positions.
\Cref{fig:deepseek-7b-probe-analysis}a shows representative results for DeepSeek-7B on the HARP dataset (similar results for other models in \Cref{app:more-probe-analysis-results}).
Probes trained on individual layers achieve similar accuracy to one another, indicating that importance is not localized to specific layers.
Moreover, no single-layer or single-position probe matches the accuracy of the full probe, which pools information across all layers and positions.
This indicates that the features underlying importance are distributed throughout the activations: individual layers capture substantial signal, but combining them is necessary to correctly classify the most ambiguous reasoning steps. 

\paragraph{Importance is not a function of surface-level features.}
We verify that the probe does not only recover surface-level features, which we define as those derivable from tokens.
We evaluate a set of candidates: 
relative position of the step within the reasoning chain, step length in tokens, number of numerical and arithmetic tokens, redundancy against previous steps, relevance to the input problem, and features that check whether specific categories of expressions appear in the step (e.g. ``Let's think.'' indicates the ``filler'' category; see \Cref{app:surface-level-features} for the full list).
Individually, all features show low correlation with the probe's predictions (\Cref{fig:deepseek-7b-probe-analysis}b).
Combining these features, a linear regression model trained on the full feature set performs better than random, but substantially worse than the activation-based probe (\Cref{fig:deepseek-7b-probe-analysis}c).
Similar results for other models are shown in \Cref{app:more-probe-analysis-results}.
This suggests that while surface-level features can help in identifying importance, our activation-based probe identifies a richer notion of importance that accurately predicts more steps.

Together, these results suggest that the probe captures a notion of importance that is distributed across the model and irreducible to surface-level token statistics.
This further indicates that full-context, full-layers latent activations are a better representation to study reasoning importance, and potentially other latent notions.

\section{Related work}
\label{sec:related-work}

\paragraph{Interpreting reasoning chains.} 
Prior work has studied reasoning chains from multiple perspectives, mostly relying on tokens: categorizing steps by relevance, coherence, and correctness \citep{jacovi2024chain, do2025definesgoodreasoningllms}; analyzing reasoning structure as a graph of semantically-clustered steps \citep{teng2025atom, feng2025characterizes, jiang2025makes}; or identifying influential steps by sampling replacements \citep{bogdan2025thoughtanchors}.
However, analysis relying on tokens has well-known limitations: generated reasoning does not always reflect model internals \citep{agarwal2024faithfulness, stechly2025unreasonable}, and many steps are ``performative'' and have minimal effect on the final answer \citep{zhang2025reasoning, zhao2025can, boppana2026reasoning}.
This limitation led to studies examining reasoning through model latents directly, via the logit lens \citep{nostalgebraist2020logit, chen2026think}, algorithmic primitives \citep{lippl2026algorithmic}, or geometric patterns \citep{zhou2025geometry, minegishi2025topology}.
Where these studies either predict correctness based on perceived thinking depth or offer high level perspectives of latent reasoning, we study the concept of importance of reasoning steps and whether it can be decoded from tokens or model latents.

\paragraph{Probing for abstract concepts.}
Probing classifiers \citep{alain2016understanding, belinkov2022probing} have been used to identify abstract concepts encoded in language model representations, including downstream performance \citep{afzal2025knowing, zhu2022predicting}, truthfulness \citep{Joshi2023PersonasAA, simhi2024constructing, orgad2024llms}, and model beliefs \citep{bortoletto2024brittle} and attributes \citep{arditi2024refusal, papadatos2024linear}.
While these probes are often linear and targeted at specific positions, cross-layer probes have been shown to yield strong results \citep{bar2025beyond}. 
Building on these, we train probes to identify step importance in reasoning chains, revealing that this abstract notion is encoded in model activations even prior to generation of subsequent steps.

\paragraph{Importance estimation for reasoning chain pruning.}
Measuring reasoning step importance can be used for pruning reasoning chains and obtaining a minimal sufficient reasoning trace \citep{ding2024break}.
Along this line, several studies prune reasoning chain in place, removing tokens or entire steps by using signals from step entropy or perplexity \citep{cui2025stepwise, li2025selfentropy, xiao2025limopro}, token attribution \citep{wang2026segment}, prompting  \citep{liu2024skip}, or external verifiers \citep{lin2025trimr}.
While these studies implicitly raise the question of \textit{``which reasoning steps are important?''}, we do so explicitly: our focus is not pruning itself, but rather using step removability as a causal intervention to investigate if and how LLMs encode the concept of step importance.

\section{Conclusions, discussion and limitations}
Do reasoning models ``know'' when a reasoning step is important?
We approached this question through the lens of removability---whether a step can be removed while still leaving an attainable path to the correct answer.
Our findings show that most reasoning steps are not important, and that model internals contain signals that accurately identify the most important steps.
These signals are best recoverable from full-context latent activations. They are distributed across layers and positions, and transferable across models.

Our results have several implications.
First, model activations encode abstract concepts that might not surface at the token level.
This suggests a form of unfaithfulness, where steps that appear significant in the generated trace are in fact removable. 
We thus argue that studying reasoning through full-context model activations offers a more faithful lens on the reasoning process.
Second, if a small subsequence of steps is both sufficient for reaching the correct answer and highly attainable, then this core reasoning subsequence---rather than the full trace---may be a more informative target for monitoring and studying internal computation.

Several of our analyses yielded negative yet worthwhile results. 
Our attempt at analyzing the importance probe weights with standard tools \citep[e.g., logit lens;][]{nostalgebraist2020logit} produced null findings; 
attempts to taxonomize core reasoning steps by their functionality were similarly inconclusive.
We further discuss these negative results in \Cref{app:negative-results}.

Our work has several limitations. Our analysis focuses on mathematical reasoning, which is a standard benchmark for evaluating reasoning models.
Extending this analysis to other domains where reasoning models excel, such as coding or open-ended problems, and validating our findings hold there, is a natural direction for further work.
Finally, while we demonstrate that models encode a concept of importance in their activations, we do not use this signal to fine-tune models on core reasoning chains; this is beyond the analytic focus of this work.
Nonetheless, grounded in our notion of attainability and in line with the pruning literature (\Cref{sec:related-work}), we believe such fine-tuning is a viable and promising direction.

\section*{Acknowledgments}
We thank Andrew Lee and Dana Arad for providing feedback on a draft of this work.
This research was supported by Coefficient Giving, the Israel Science Foundation (grant No. 2942/25), and the European Union (ERC, Control-LM, 101165402).
Views and opinions expressed are however those of the author(s) only and do not necessarily reflect those of the European Union or the European Research Council Executive Agency. Neither the European Union nor the granting authority can be held responsible for them.
This work was partly supported by a Gemini Academic Program Award.
YN was supported by the Council for Higher Education (VATAT) Scholarship for PhD students in data science and artificial intelligence.

\bibliography{colm2026_conference}
\bibliographystyle{colm2026_conference}

\newpage
\appendix

\crefalias{section}{appendix} %
\crefalias{subsection}{appendix} %
\crefalias{subsubsection}{appendix} %

\section{Discussion on removability}
\label{app:removability-discussion}

\paragraph{Motivation for using step removability as a proxy for importance.}
We investigate whether LLMs latently encode a notion of reasoning step importance.
To operationalize importance, we adopt a causal perspective: a step is important to the extent that intervening on it affects the model's ability to process the reasoning chain and produce a correct answer.
The first practical choice is the granularity of the intervention: intervening on a single token or a single step in isolation is often not enough to induce a measurable effect (see \Cref{app:negative-results}), so we operate at the level of reasoning subsequences.
The second choice is the nature of the intervention itself. 
One can remove a step, substitute it with a paraphrase, an uninformative filler, or an incorrect alternative \citep{lanham2023measuring}.
However, applying substitution-based interventions across multiple steps is problematic.
It risks entangling the intervention with uncontrolled confounders of the replacement process, making it difficult to attribute downstream behavioral changes solely to targeted steps.
Further more, this introduces generative overhead, requiring a capable model to perform consistent replacements.
We therefore adopt step removal: it is simple to apply in parallel across multiple steps and has a direct effect on downstream reasoning.
Furthermore, it can have practical implications for reasoning efficiency: if removability can be accurately detected from latent activations, the probes that detect it could be used at inference time to omit removable steps and reduce context length.

\paragraph{Alternative definition of removable steps.} 
In \Cref{sec:method}, we define removable steps as those removed during the first stage of identifying the core reasoning subsequence $\mathcal{T}^*$.
An alternative would be to treat any step not in $\mathcal{T}^*$ as removable, additionally including steps eliminated during the second stage ($\mathcal{T}_{\mathrm{inf}} \to \mathcal{T}^*$).
We opt against this broader definition because step importance is not binary but continuous: steps removed during the first stage occupy the low-importance extreme, steps in $\mathcal{T}^*$ occupy the high-importance extreme, and the remaining steps fall somewhere in between.
We restrict our analysis to steps at these two extremes. This yields the cleanest ``ground-truth'' labels for our classification experiments in \Cref{sec:identifying-importance}.

\section{Technical details}
\label{app:technical-details}

We provide technical details for each part in the experimentation process.

\subsection{Reasoning chain generation, step segmentation and evaluation}
\label{app:technical-details-generation-segmentation-evaluation}
\paragraph{Reasoning chain generation.}
For each input prompt $x$, we generate a single deterministic reasoning chain of up to $8,000$ tokens.
This limit is imposed for several reasons. 
First, to maintain efficiency, given the high volume of analyzed prompts (See \Cref{app:performance-stats}).
Second, we found that across models, increasing the maximal chain size only led to a slight increase in model accuracy. Thus we chose this limit as a compromise. 
We use greedy decoding as it leads to sufficient model performance on the reasoning datasets and alleviates the need to validate sampling hyper-parameters.
For the HARP dataset, which is divided into $6$ difficulty levels, we analyze a range of  difficulties fit to each model's capability.
The two strongest models, GPT-OSS-20B and DeepSeek-R1-Qwen-14B, are evaluated across all difficulty levels. 
DeepSeek-R1-Qwen-1.5B, DeepSeek-R1-Qwen-7B and Olmo-3-7B-Think are evaluated on difficulties 1–4.
For the MATH-500 dataset, all $500$ prompts are analyzed across all models.
For GPT-OSS-20B, we use the default ``medium'' reasoning depth setting.

\paragraph{Segmentation into reasoning steps.}
To segment a reasoning chain into steps, we follow standard procedure \citep{bogdan2025thoughtanchors, zhang2025reasoning, minegishi2025topology}, splitting at paragraph-ending patterns (two consecutive newline characters) or sentence-ending patterns (a period, question mark, or exclamation mark followed by a space or newline).

\paragraph{Evaluation.}
When forcing a model to generate an answer after a given reasoning subsequence (as part of sufficiency evaluation), we append a model-specific suffix $\sigma$ to elicit the final response.
For DeepSeek and Olmo models, this takes the form:
\begin{equation}
    \sigma = \texttt{``So the answer is }\backslash \texttt{boxed\{''}\ ,
\end{equation}
whereas for GPT-OSS we use:
\begin{equation}
    \sigma = \texttt{``<|end|><|start|>final<|message|>The answer is }\backslash\texttt{boxed\{''}\ .
\end{equation}
We chose these specific suffixes to best match the way these models generate a final answer during standard generation. 

To evaluate sufficiency and necessity, we sample $5$ different completions from the model.
We use a sampling temperature of $0.6$ and $\text{top\_p}=0.95$. 

Throughout the analysis, we use fixed thresholds of $\mathrm{Sufficiency} = 0.8$ for sufficiency, $\mathrm{Attainability} = 0.8$ for attainability (see discussion and ablations in \Cref{app:suff-atnb-threshold-ablation}).

\paragraph{Answer comparisons.}
In mathematical reasoning, the same answer can be represented in different ways.
For instance, the answer $-10 \le x \le 1$ is equivalent to $x \in [-10, 1]$; just as $5 + x$ and $x+5$ are identical expressions.
Thus, when a model completes a given answer, the comparison to the ground truth answer must take these representations into account.
To account for these cases, we compare generated answers against ground truth using a suite of heuristics designed to handle the variety of forms in which the results may appear, including tuples, ranges, fractions, numeric quantities, and symbolic expressions.
Full implementation details are available in our code.
It is also noteworthy that this discrepancy limits us from directly using token probabilities in our analysis.

\paragraph{Eliminating shortcut steps.}
Reasoning chains often contain steps that, when viewed in isolation, can act as potential shortcuts to the final answer.
Such shortcuts can harm the validity of our sufficiency measure for any reasoning chain that contains them.
To prevent these shortcuts from confounding our analysis, we remove these steps from the generated chains prior to the identification of a sufficient prefix (\Cref{sec:prelims}) and core reasoning subsequence (\Cref{sec:method}).
To identify shortcut steps, we measure the sufficiency of each individual step separately (i.e. we measure $\mathrm{Sufficiency}(\mathcal{T})$ when $\mathcal{T}=(s_i)$ for each step $s_i$).
We consider a step to be a shortcut step if it answers at least one of the following criteria:
\begin{itemize}
    \item it alone is sufficient to reach the answer (i.e. $\mathrm{Sufficiency}((s_i))\ge\tau$);
    \item it contains a boxed answer expression (i.e. ``\texttt{\textbackslash boxed\{\ldots\}}'') whose value is equivalent to the ground-truth answer;
    \item it contains a direct answer statement (a phrase beginning with one of ``$\texttt{Thus}$'', ``$\texttt{Therefore}$'', ``$\texttt{So}$'', ``$\texttt{Hence}$'', ``$\texttt{Consequently}$'', ``$\texttt{This means}$'', ``$\texttt{My}$'',  ``$\texttt{The}$''; contains the word ``$\texttt{answer}$'' or ``$\texttt{result}$''; and some answer value).
\end{itemize}

\subsection{Identifying a core reasoning subsequence}
\label{app:labeling-importance}
In this section we present the full details on the different variants to identify the core reasoning subsequence ($\mathcal{T}^*$), as presented in \Cref{sec:method}. 

\subsubsection{Activation-based variant}
\label{app:activation-based-labeling-procedure}
We present the full code for the activation-based variant.
Function~\ref{alg:activation-variant-first-stage} implements the first stage, building the sequence of steps with direct or indirect influence on the answer ($\mathcal{T}_{\mathrm{inf}}$).
Steps excluded from $\mathcal{T}_{\mathrm{inf}}$ are labeled as removable for later classification experiments.

Function~\ref{alg:activation-variant-second-stage} implements the second stage, building the core reasoning subsequence $\mathcal{T}^*$.
Steps included in $\mathcal{T}^*$ are labeled as important and non-removable.

\begin{lstlisting}[caption={Activation-based variant function to build $\mathcal{T}_{\mathrm{inf}}$.}, label={alg:activation-variant-first-stage},float,
                   floatplacement=H]
def get_influence_mask(influence_matrix, threshold):
  '''
  Args:
  - influence_matrix: A (p x p) matrix where entry (j, i) represents the 
      influence of step i on step j.
  - threshold: A float between 0 and 1 representing the fraction of total
      influence to retain for each step.
  '''
  p = influence_matrix.shape[0]
  step_mask = torch.zeros(p, dtype=torch.long)
  step_mask[p - 1] = 1
  for ref_idx in range(p - 1, 0, -1):
    if step_mask[ref_idx] != 0:
      influence_on_step = influence_matrix[ref_idx, :]
      
      # Sort in descending order
      sorted_inf = torch.sort(influence_on_step, descending=True).values
      
      # Accumulate and Normalize
      cumulative_score = torch.cumsum(sorted_inf, dim=0) / torch.sum(sorted_inf)
      
      # Find score threshold that keeps the desired fraction of total influence
      threshold_index = torch.searchsorted(cumulative_score, threshold).item()
      threshold_influence_value = sorted_inf[threshold_index]
      
      # Include all steps that contribute to the influence in the mask
      step_mask |= (influence_on_step >= threshold_influence_value)
      
  return step_mask
\end{lstlisting}

\begin{lstlisting}[caption={Activation-based variant function to build $\mathcal{T}^*$.}, label={alg:activation-variant-second-stage},float,
                   floatplacement=H]
def get_core_reasoning_subsequence(influence_matrix):
  '''
  Args:
  - influence_matrix: A (p x p) matrix where entry (j, i) represents the 
      influence of step i on step j.
  '''
  THRESHOLDS = [0.1, 0.2, 0.3, 0.4, 0.5, 0.6, 0.7, 0.8, 0.9, 1.0]
  for t in THRESHOLDS:
    mask = get_influence_mask(influence_matrix, t)

    suff, atnb = evaluate_reasoning_subsequence(mask)
    # Stop if sufficient mask is found or all steps are kept
    if suff >= SUFFICIENCY_THRESHOLD and atnb >= ATTAINABILITY_THRESHOLD:
      break

  # Calculate average influence
  mean_influence = influence_matrix.sum(dim=0) / (influence_matrix != 0).sum(dim=0)

  # Get ranking by average influence
  ranked_step_indices = mean_influence.argsort()

  # Only consider steps that were kept in the last threshold result
  # Build queue sorted by ascending mean influence
  queue = [idx for idx in mean_influence.argsort() 
           if mask[idx] == 1 and idx != len(mask) - 1]
  failures_since_last_removal = 0
  
  while queue and failures_since_last_removal < len(queue):
    step_idx = queue.pop(0)
    eval_mask = mask.clone()
    eval_mask[step_idx] = 0
    suff, atnb = evaluate_reasoning_subsequence(eval_mask)
    if suff >= SUFFICIENCY_THRESHOLD and atnb >= ATTAINABILITY_THRESHOLD:
      # Confirm removal
      mask = eval_mask
      failures_since_last_removal = 0
    else:
      # Revert and re-queue at end
      queue.append(step_idx)
      failures_since_last_removal += 1
  
  core_subsequence = [step_idx for step_idx in range(len(mask)) if mask[step_idx] == 1]
  return core_subsequence
\end{lstlisting}

\subsubsection{Token-based variant}
\label{app:token-based-labeling-procedure}
For the token-based variant, we use Gemini-2.5-Pro \citep{comanici2025gemini} as the external LLM for both stages of the greedy procedure ($\mathcal{T}_p\to\mathcal{T}_{\mathrm{inf}}$ and $\mathcal{T}_{\mathrm{inf}} \to \mathcal{T}^*$). 
In both stages, Gemini is called through the Gemini API, using greedy decoding (temperature $t = 0$) with up to $24k$ output tokens. 
We approximate the total cost of API calls for this at $50$ USD. 

In the first stage, we prompt Gemini to identify reasoning steps that do not contribute to the final answer (Prompt~\ref{prompt:token-based-variant-first-stage}); steps excluded from the result are labeled as removable.
We then prompt Gemini for a ranking of the remaining steps by importance (Prompt~\ref{prompt:token-based-variant-second-stage}).
This ranking drives the iterative removal of the least important steps (similarly to \Cref{alg:activation-variant-second-stage}), and results in the core reasoning subsequence $\mathcal{T}^*$. 

\begin{listing}[H]
\begin{mdframed}[
  linewidth=1pt,
  nobreak=true,
  linecolor=black,
  backgroundcolor=gray!10,
  innertopmargin=10pt,
  innerbottommargin=10pt,
  innerrightmargin=15pt,
  innerleftmargin=15pt
]
You are analyzing a mathematical chain-of-thought (CoT) reasoning trace.

You are given the input question and the minimal reasoning trace required to answer, split to steps.

Your task is to identify which reasoning steps do NOT contribute directly or indirectly to producing the final answer.

A step contributes directly if it is used in reaching the final answer.

A step contributes indirectly if it influences another step that recursively affects the final answer.

A step does NOT contribute if removing it would not affect the reasoning chain's  ability to reach the correct answer — for example, steps that are redundant, restatements, or tangential.

Don't consider the last sentence (the answer step).

The question: $<$question$>$\{question\}$<$/question$>$

The reasoning steps:

\{sentence\_lines\}

After thinking, output your final answer as a JSON list of step indices that do NOT contribute to the answer and can be safely removed. Output the list inside a ```json code block, like:

```json

$[\text{removable\_idx\_1, removable\_idx\_2, ...}]$

```
\end{mdframed}
\caption{Gemini prompt to identify $\mathcal{T}_{\mathrm{inf}}$ in the token-based variant.}
\label{prompt:token-based-variant-first-stage}
\end{listing}

\begin{listing}[H]
\begin{mdframed}[
  linewidth=1pt,
  nobreak=true,
  linecolor=black,
  backgroundcolor=gray!10,
  innertopmargin=10pt,
  innerbottommargin=10pt,
  innerrightmargin=15pt,
  innerleftmargin=15pt
]
You are analyzing a mathematical chain-of-thought (CoT) reasoning trace.

You are given the input question and the minimal reasoning trace required to answer, split to steps.

Your task is to rank ALL reasoning steps by their importance in producing the final answer.

A step is important if it directly contributes to the final answer, or if it indirectly influences another step that recursively affects the final answer. A step is unimportant if removing it would not affect the reasoning chain's ability to reach the correct answer.

Think about which steps are load-bearing for the argument and which are redundant, restatements, or tangential. 

Don't rank the last sentence.

The question: $<$question$>$\{question\}$<$/question$>$

The reasoning steps:

\{sentence\_lines\}

After thinking, output your final ranking as a JSON list of step indices ordered from MOST important to LEAST important. The list must contain exactly the indices \{valid\_indices\}. 

Output the ranking inside a ```json code block, like:

```json

$\text{[most\_important\_idx, ..., least\_important\_idx]}$

```
\end{mdframed}
\caption{Gemini prompt to rank $\mathcal{T}_{\mathrm{inf}}$ steps by importance for the token-based variant.}
\label{prompt:token-based-variant-second-stage}
\end{listing}

\subsection{Classifying reasoning steps}
\label{app:classification-reasoning-steps}

In this section we present the full details on methods for importance classification (\Cref{sec:identifying-importance}).

\subsubsection{Activation-based probe importance classification}
\label{app:importance-classifier-technical-details}

In this section we describe the technical details regarding the activation-based importance probes.

\paragraph{Probe architectures and input representations.} We evaluate four probe families:
\begin{enumerate}
    \item Linear: a single linear layer from the averaged activation vector to a scalar.
    \item MLP: an MLP with a ReLU non-linearity. We sweep over hidden dimensions $128, 256$ and depth $2, 5$.
    \item Layerwise-Linear (LW-Linear): We flatten the $(L, D)$ per-layer activation representation and apply a single linear layer.
    \item Layerwise-MLP (LW-MLP): the same flattened input is passed through a $2$-layer MLP. We sweep over the hidden probe dimension from ${16, 128, 1024, 4096}$.
\end{enumerate}
In the first two architectures (Linear and MLP), the activation tensors are averaged across both token and layer dimensions, resulting in $d-$shaped inputs, where $d$ is the model's hidden dimension.
Across the latter architectures (LW-Linear and LW-MLP), the activations are averaged only across the token dimension, resulting in a $(L,d)-$shaped inputs.
We additionally tested shallow transformer encoders ($1$--$2$ layers) operating on the full $(S,L,d)-$shaped activation tensors, but these underperformed the simpler probes while incurring substantially higher training costs.

\paragraph{Training.} Probes are trained for $30$ epochs with AdamW (with learning rate $3\times10^{-4}$, weight decay $10^{-4}$, batch size $32$ and cosine annealing learning rate scheduler) using binary cross-entropy loss. 
The labeled reasoning steps are split to an $80/20$ training/evaluation split, and repeat the process across $3$ seeds.
We report the best evaluation accuracy achieved across epochs.

\paragraph{Results.} The results across all tested settings are presented in \Cref{tab:probe-arch-search-harp,tab:probe-arch-search-math500} for the HARP and MATH-500 datasets, respectively.
Across most model–dataset combinations, the layerwise-MLP probe with hidden dimension $128$ performs best, motivating our choice of it as the standard probe in \Cref{sec:activation-importance-exp,sec:understanding-importance}.

\begin{table}[htbp]
\centering
\resizebox{\textwidth}{!}{
\begin{tabular}{lccccc}
\toprule
\textbf{Configuration} & \textbf{GPT-OSS} & \textbf{DS-1.5} & \textbf{DS-7} & \textbf{DS-14} & \textbf{Olmo3} \\
\midrule
  Context, All Layers, Linear & 73.0\%$_{\pm 0.6\%}$ & 75.6\%$_{\pm 1.0\%}$ & 77.8\%$_{\pm 1.0\%}$ & 77.6\%$_{\pm 0.3\%}$ & 75.0\%$_{\pm 0.6\%}$ \\
  Context, All Layers, MLP (h=128, L=2) & 79.2\%$_{\pm 0.1\%}$ & 78.8\%$_{\pm 1.1\%}$ & 81.5\%$_{\pm 0.5\%}$ & 80.3\%$_{\pm 0.2\%}$ & 80.3\%$_{\pm 0.2\%}$ \\
  Context, All Layers, MLP (h=128, L=5) & 78.6\%$_{\pm 0.4\%}$ & 78.3\%$_{\pm 1.3\%}$ & 81.3\%$_{\pm 0.9\%}$ & 80.6\%$_{\pm 0.7\%}$ & 79.6\%$_{\pm 0.3\%}$ \\
  Context, All Layers, MLP (h=256, L=2) & 79.1\%$_{\pm 0.3\%}$ & 79.4\%$_{\pm 1.6\%}$ & 82.0\%$_{\pm 1.1\%}$ & 81.0\%$_{\pm 0.3\%}$ & 80.2\%$_{\pm 0.4\%}$ \\
  Context, All Layers, MLP (h=256, L=5) & 79.1\%$_{\pm 0.2\%}$ & 79.4\%$_{\pm 1.7\%}$ & 81.8\%$_{\pm 1.0\%}$ & 80.9\%$_{\pm 0.4\%}$ & 80.1\%$_{\pm 0.2\%}$ \\
  Context, All Layers, LW-Linear & 75.0\%$_{\pm 0.4\%}$ & 77.6\%$_{\pm 1.0\%}$ & 80.4\%$_{\pm 0.7\%}$ & 79.7\%$_{\pm 0.2\%}$ & 79.2\%$_{\pm 0.3\%}$ \\
  Context, All Layers, LW-MLP (h=16) & 78.5\%$_{\pm 0.3\%}$ & 79.5\%$_{\pm 1.3\%}$ & 82.4\%$_{\pm 0.3\%}$ & 81.9\%$_{\pm 0.5\%}$ & 80.5\%$_{\pm 0.6\%}$ \\
  Context, All Layers, LW-MLP (h=128) & \textbf{80.0\%}$_{\pm 0.2\%}$ & \textbf{80.6\%}$_{\pm 1.1\%}$ & 82.5\%$_{\pm 0.2\%}$ & 82.5\%$_{\pm 0.3\%}$ & \textbf{81.6\%}$_{\pm 0.6\%}$ \\
  Context, All Layers, LW-MLP (h=1024) & \textbf{80.0\%}$_{\pm 0.3\%}$ & 80.3\%$_{\pm 1.2\%}$ & \textbf{82.6\%}$_{\pm 0.3\%}$ & 82.4\%$_{\pm 0.4\%}$ & 81.1\%$_{\pm 0.8\%}$ \\
  Context, All Layers, LW-MLP (h=4096) & 80.0\%$_{\pm 0.1\%}$ & 80.0\%$_{\pm 1.2\%}$ & 82.4\%$_{\pm 0.2\%}$ & \textbf{82.6\%}$_{\pm 0.2\%}$ & 81.5\%$_{\pm 0.6\%}$ \\
  No Context, All Layers, Linear & 66.6\%$_{\pm 0.2\%}$ & 56.7\%$_{\pm 0.8\%}$ & 57.6\%$_{\pm 1.1\%}$ & 71.0\%$_{\pm 0.7\%}$ & 70.4\%$_{\pm 0.6\%}$ \\
  No Context, All Layers, MLP (h=128, L=2) & 69.2\%$_{\pm 0.2\%}$ & 56.8\%$_{\pm 0.9\%}$ & 57.9\%$_{\pm 0.8\%}$ & 72.4\%$_{\pm 1.1\%}$ & 70.4\%$_{\pm 0.2\%}$ \\
  No Context, All Layers, MLP (h=128, L=5) & 69.4\%$_{\pm 0.4\%}$ & 56.9\%$_{\pm 0.9\%}$ & 56.9\%$_{\pm 0.6\%}$ & 71.3\%$_{\pm 0.0\%}$ & 71.3\%$_{\pm 0.4\%}$ \\
  No Context, All Layers, MLP (h=256, L=2) & 69.0\%$_{\pm 0.0\%}$ & 56.7\%$_{\pm 0.9\%}$ & 56.9\%$_{\pm 0.5\%}$ & 72.9\%$_{\pm 0.8\%}$ & 70.4\%$_{\pm 0.1\%}$ \\
  No Context, All Layers, MLP (h=256, L=5) & 68.8\%$_{\pm 0.4\%}$ & 56.7\%$_{\pm 0.9\%}$ & 57.7\%$_{\pm 0.1\%}$ & 71.8\%$_{\pm 0.6\%}$ & 70.7\%$_{\pm 0.1\%}$ \\
  No Context, All Layers, LW-Linear & 65.5\%$_{\pm 0.2\%}$ & 56.9\%$_{\pm 0.8\%}$ & 56.7\%$_{\pm 0.2\%}$ & 71.0\%$_{\pm 0.5\%}$ & 71.1\%$_{\pm 0.4\%}$ \\
  No Context, All Layers, LW-MLP (h=16) & 69.0\%$_{\pm 0.5\%}$ & 56.7\%$_{\pm 0.9\%}$ & 58.3\%$_{\pm 0.7\%}$ & 72.5\%$_{\pm 0.4\%}$ & 70.7\%$_{\pm 0.4\%}$ \\
  No Context, All Layers, LW-MLP (h=128) & 68.9\%$_{\pm 0.3\%}$ & 56.7\%$_{\pm 0.9\%}$ & 57.3\%$_{\pm 0.1\%}$ & 72.7\%$_{\pm 0.5\%}$ & 70.5\%$_{\pm 0.6\%}$ \\
  No Context, All Layers, LW-MLP (h=1024) & 69.6\%$_{\pm 0.2\%}$ & 56.7\%$_{\pm 0.9\%}$ & 57.7\%$_{\pm 0.6\%}$ & 71.4\%$_{\pm 0.2\%}$ & 70.9\%$_{\pm 0.2\%}$ \\
  No Context, All Layers, LW-MLP (h=4096) & 69.6\%$_{\pm 0.3\%}$ & 56.7\%$_{\pm 0.9\%}$ & 57.7\%$_{\pm 1.0\%}$ & 71.9\%$_{\pm 0.3\%}$ & 71.4\%$_{\pm 0.3\%}$ \\
\bottomrule
\end{tabular}}
\caption{Probe architecture search results on HARP}
\label{tab:probe-arch-search-harp}
\end{table}

\begin{table}[htbp]
\centering
\resizebox{\textwidth}{!}{
\begin{tabular}{lccccc}
\toprule
\textbf{Configuration} & \textbf{GPT-OSS} & \textbf{DS-1.5} & \textbf{DS-7} & \textbf{DS-14} & \textbf{Olmo3} \\
\midrule
  Context, All Layers, Linear & 78.0\%$_{\pm 1.2\%}$ & 75.8\%$_{\pm 3.8\%}$ & 82.9\%$_{\pm 0.9\%}$ & 74.0\%$_{\pm 3.1\%}$ & 78.3\%$_{\pm 0.5\%}$ \\
  Context, All Layers, MLP (h=128, L=2) & 83.4\%$_{\pm 1.3\%}$ & 77.8\%$_{\pm 1.8\%}$ & \textbf{86.1\%}$_{\pm 1.5\%}$ & 79.2\%$_{\pm 1.7\%}$ & 82.0\%$_{\pm 1.5\%}$ \\
  Context, All Layers, MLP (h=128, L=5) & 82.3\%$_{\pm 1.5\%}$ & 72.6\%$_{\pm 3.4\%}$ & 85.9\%$_{\pm 1.3\%}$ & 71.9\%$_{\pm 2.3\%}$ & 82.4\%$_{\pm 0.8\%}$ \\
  Context, All Layers, MLP (h=256, L=2) & 83.3\%$_{\pm 0.9\%}$ & 78.3\%$_{\pm 1.0\%}$ & 86.0\%$_{\pm 1.8\%}$ & 78.5\%$_{\pm 1.3\%}$ & 82.3\%$_{\pm 1.9\%}$ \\
  Context, All Layers, MLP (h=256, L=5) & 83.3\%$_{\pm 1.0\%}$ & 77.1\%$_{\pm 1.5\%}$ & 84.8\%$_{\pm 1.6\%}$ & 75.0\%$_{\pm 2.3\%}$ & \textbf{82.8\%}$_{\pm 1.3\%}$ \\
  Context, All Layers, LW-Linear & 80.9\%$_{\pm 1.5\%}$ & 80.5\%$_{\pm 3.7\%}$ & 85.6\%$_{\pm 1.7\%}$ & 79.5\%$_{\pm 2.7\%}$ & 80.5\%$_{\pm 0.9\%}$ \\
  Context, All Layers, LW-MLP (h=16) & 83.4\%$_{\pm 1.7\%}$ & \textbf{81.3\%}$_{\pm 0.6\%}$ & 82.3\%$_{\pm 2.5\%}$ & 78.5\%$_{\pm 2.7\%}$ & 82.3\%$_{\pm 1.7\%}$ \\
  Context, All Layers, LW-MLP (h=128) & 83.8\%$_{\pm 1.0\%}$ & 80.5\%$_{\pm 0.9\%}$ & 83.5\%$_{\pm 1.9\%}$ & \textbf{80.5\%}$_{\pm 2.2\%}$ & \textbf{82.8\%}$_{\pm 1.9\%}$ \\
  Context, All Layers, LW-MLP (h=1024) & 83.4\%$_{\pm 1.1\%}$ & 80.4\%$_{\pm 1.6\%}$ & 83.0\%$_{\pm 1.8\%}$ & 79.9\%$_{\pm 4.8\%}$ & \textbf{82.8\%}$_{\pm 1.8\%}$ \\
  Context, All Layers, LW-MLP (h=4096) & \textbf{84.2\%}$_{\pm 0.4\%}$ & 79.4\%$_{\pm 1.0\%}$ & 83.1\%$_{\pm 2.8\%}$ & 77.8\%$_{\pm 5.8\%}$ & 82.2\%$_{\pm 1.6\%}$ \\
  No Context, All Layers, Linear & 72.0\%$_{\pm 0.4\%}$ & 58.5\%$_{\pm 2.1\%}$ & 58.5\%$_{\pm 2.1\%}$ & 58.0\%$_{\pm 2.7\%}$ & 74.5\%$_{\pm 1.5\%}$ \\
  No Context, All Layers, MLP (h=128, L=2) & 75.2\%$_{\pm 0.0\%}$ & 60.5\%$_{\pm 1.5\%}$ & 58.9\%$_{\pm 1.0\%}$ & 61.1\%$_{\pm 1.3\%}$ & 73.6\%$_{\pm 1.0\%}$ \\
  No Context, All Layers, MLP (h=128, L=5) & 73.7\%$_{\pm 1.5\%}$ & 60.5\%$_{\pm 2.4\%}$ & 59.6\%$_{\pm 1.7\%}$ & 56.9\%$_{\pm 3.5\%}$ & 73.6\%$_{\pm 0.3\%}$ \\
  No Context, All Layers, MLP (h=256, L=2) & 75.5\%$_{\pm 0.9\%}$ & 59.4\%$_{\pm 0.8\%}$ & 59.0\%$_{\pm 1.4\%}$ & 61.1\%$_{\pm 2.1\%}$ & 73.6\%$_{\pm 0.5\%}$ \\
  No Context, All Layers, MLP (h=256, L=5) & 75.2\%$_{\pm 0.7\%}$ & 59.4\%$_{\pm 1.6\%}$ & 57.3\%$_{\pm 2.0\%}$ & 59.7\%$_{\pm 4.0\%}$ & 73.4\%$_{\pm 0.3\%}$ \\
  No Context, All Layers, LW-Linear & 73.4\%$_{\pm 0.6\%}$ & 59.7\%$_{\pm 2.6\%}$ & 58.9\%$_{\pm 0.7\%}$ & 59.0\%$_{\pm 4.7\%}$ & 73.0\%$_{\pm 0.5\%}$ \\
  No Context, All Layers, LW-MLP (h=16) & 75.8\%$_{\pm 1.1\%}$ & 59.4\%$_{\pm 1.0\%}$ & 59.6\%$_{\pm 0.9\%}$ & 59.4\%$_{\pm 1.5\%}$ & 72.6\%$_{\pm 1.4\%}$ \\
  No Context, All Layers, LW-MLP (h=128) & 75.6\%$_{\pm 2.0\%}$ & 57.6\%$_{\pm 2.6\%}$ & 60.3\%$_{\pm 1.1\%}$ & 61.5\%$_{\pm 1.5\%}$ & 72.3\%$_{\pm 0.6\%}$ \\
  No Context, All Layers, LW-MLP (h=1024) & 74.0\%$_{\pm 1.5\%}$ & 59.3\%$_{\pm 1.9\%}$ & 59.7\%$_{\pm 2.0\%}$ & 61.8\%$_{\pm 3.4\%}$ & 74.3\%$_{\pm 0.4\%}$ \\
  No Context, All Layers, LW-MLP (h=4096) & 75.3\%$_{\pm 0.4\%}$ & 59.4\%$_{\pm 1.0\%}$ & 57.6\%$_{\pm 1.5\%}$ & 61.1\%$_{\pm 1.3\%}$ & 74.1\%$_{\pm 0.4\%}$ \\
\bottomrule
\end{tabular}}
\caption{Probe architecture search results on MATH-500}
\label{tab:probe-arch-search-math500}
\end{table}

\subsubsection{Baselines for importance classification}
\label{app:baselines-for-importance-classification}
In this section we describe the technical details regarding the baselines for importance detection (\Cref{sec:baselines-for-importance-detection}).

\paragraph{LLM-as-a-judge.} We prompt Gemini-2.5-Pro \citep{comanici2025gemini} with a classification prompt containing a target reasoning step, asking it to determine whether the step is removable or non-removable.
We use greedy decoding (temperature $t = 0$) via the Gemini API, allowing up to $24k$ output tokens. 
The number of labeled reasoning steps can reach tens of thousands of entries, making classification prohibitively expensive. 
Thus, we sample $1500$ steps per model per dataset (while maintaining a balanced split between removable and non-removable labels).
We approximate the total cost of API calls for this at $150$ USD.
The classification prompts are provided in Prompts~\ref{prompt:llm-as-a-judge-no-context-classification} and \ref{prompt:llm-as-a-judge-full-context-classification} for classification without context and with full context, respectively.

\paragraph{TF-IDF representation baseline.} 
We represent each reasoning step as a sparse TF-IDF vector using scikit-learn's \citep{scikit-learn} \texttt{TfidfVectorizer} with sublinear term-frequency scaling. We use $n$-grams of $n \in [1, 5]$, and a vocabulary capped at $100{,}000$ features. 
We use a two-layer MLP with an input dimension of $100,000$, a hidden dimension of $128$ and a ReLU non-linearity. We train it for $30$ epochs with AdamW (with learning rate $1\times10^{-3}$, weight decay $10^{-4}$, batch size $256$) using binary cross-entropy loss.

\paragraph{Sentence transformer representation baseline.} We use \texttt{all-MiniLM-L6-v2} as the pre-trained sentence transformer.
We use the same classifier and training scheme as the TF-IDF baseline.

\begin{listing}[H]
\begin{mdframed}[
  linewidth=1pt,
  nobreak=true,
  linecolor=black,
  backgroundcolor=gray!10,
  innertopmargin=10pt,
  innerbottommargin=10pt,
  innerrightmargin=15pt,
  innerleftmargin=15pt
]
You are labeling a sentence from a mathematical chain-of-thought (CoT) reasoning trace.

Your task is to predict whether this sentence is **removable** or **nonremovable**.

Definitions:

- **removable**: The sentence can be deleted from the trace such that all subsequent sentences remain logically supported and the final answer is still reachable from the  remaining sentences alone.

- **nonremovable**: The sentence is essential to the reasoning chain. Without it, at least one subsequent sentence loses its logical basis, or the final answer can no longer be derived from the remaining sentences.

To make your judgment, consider whether the sentence:

- Introduces a new variable, equation, or fact that is used by a later sentence.

- Performs a calculation or derivation whose result is referenced downstream.

- Or instead restates the problem, expresses uncertainty, narrates the approach, summarizes prior steps, or makes a tangential observation.

Based only on the sentence text, predict whether this type of sentence is typically removable or nonremovable in a reasoning trace.

Think step by step about what role this sentence most likely plays, then on the last line output ONLY the word "removable" or "nonremovable".

Sentence: \{sentence\}
\end{mdframed}
\caption{Prompt to classify a reasoning step as removable or non-removable, without context.}
\label{prompt:llm-as-a-judge-no-context-classification}
\end{listing}

\begin{listing}[H]
\begin{mdframed}[
  linewidth=1pt,
  linecolor=black,
  nobreak=true,
  backgroundcolor=gray!10,
  innertopmargin=10pt,
  innerbottommargin=10pt,
  innerrightmargin=15pt,
  innerleftmargin=15pt
]
You are labeling a sentence from a mathematical chain-of-thought (CoT) reasoning trace.

Your task is to predict whether this sentence is **removable** or **nonremovable**.

Definitions:

- **removable**: The sentence can be deleted from the trace such that all subsequent
sentences remain logically supported and the final answer is still reachable from the
remaining sentences alone.

- **nonremovable**: The sentence is essential to the reasoning chain. Without it, at
least one subsequent sentence loses its logical basis, or the final answer can no longer be derived from the remaining sentences.

To make your judgment, consider whether the sentence:\\
- Introduces a new variable, equation, or fact that is used by a later sentence.\\
- Performs a calculation or derivation whose result is referenced downstream.\\
- Or instead restates the problem, expresses uncertainty, narrates the approach, \\
summarizes prior steps, or makes a tangential observation.

You are given the input question and the full CoT reasoning trace with sentence indices.
The target sentence to label is marked with $\texttt{>>>}$. Read the full chain of thought, then assess whether the marked sentence's content is used or referenced by any subsequent sentence, or whether it provides information necessary to reach the final answer.

Think step by step about the sentence's role in the chain, then on the last line output ONLY the word "removable" or "nonremovable".

The question: $\texttt{<question>}$\{question\}$\texttt{</question>}$

The CoT sentences (the target sentence is marked with $\texttt{>>>}$):

$[0]\ \texttt{FIRST STEP}$ \\
$[1]\ \texttt{SECOND STEP}$ \\
$ > > > [2]\ \texttt{TARGET STEP}$ \\
$[3]\ \texttt{FOURTH STEP}$ \\
$[4]\ \ldots$
\end{mdframed}
\caption{Prompt to classify a reasoning step as removable or non-removable, with full context.}
\label{prompt:llm-as-a-judge-full-context-classification}
\end{listing}

\subsection{Correlation of probes with surface-level features}
\label{app:surface-level-features}
The surface-level analysis (in \Cref{sec:concept-analysis}) tests whether the performance of activation-based importance probes can be matched by shallow, surface-level textual cues.
We extract the following surface-level features for each reasoning step:
\begin{itemize}
    \item \textbf{Relative position}: The relative position of the reasoning step, normalized by the number of reasoning steps (for step $s_i$ out of $n$ steps, this is equal to $\frac{i}{n}$.
    
    \item \textbf{Token length}: The number of tokens within the reasoning step.

    \item \textbf{Numbers count}: Count of separate numeric literals in the reasoning step.

    \item \textbf{Arithmetic operator count}: Count of arithmetic and relational symbols in $\{+,-,*,/,^,=,<,>,\leq,\geq,\neq,\pm\}$.
    
    \item \textbf{Step redundancy}: The average similarity to prior steps. We calculate the average cosine similarity between the TF-IDF embeddings of the current step $s_i$ and each preceding step in the chain, $\frac{1}{i}\sum_{j=0}^{i-1} \mathrm{sim}(s_i, s_j)$. Captures how redundant the step is with respect to earlier reasoning steps on average.

    \item \textbf{Step relevance to the input prompt}: The cosine similarity between the TF-IDF embeddings of the current reasoning step $s_i$ and the input prompt $x$. Captures how lexically related the step is to the prompt.
    
\end{itemize}

We also extract a set of categorical features, where is reasoning step is assigned a binary label for each of the following categories, set to $1$ if the step contains any of the category's phrases (matches case-insensitively). The categories and their corresponding phrases are:
         
\begin{itemize}
\item \textbf{Filler}: ``let me think'', ``let's think'', ``lets think'', ``let me consider'', ``let's consider'', ``let me see'', ``let's see'', ``let me ensure'', ``let's ensure'', ``lets ensure'', ``hmm'', ``okay'', ``alright'', ``so basically'', ``in other words''.

\item \textbf{Verification}: ``wait'', ``double check'', ``double-check'', ``recompute'', ``recalculate'', ``verify'', ``let me check'', ``let's check'', ``checking'', ``is this correct'', ``does this make sense'', ``let me re'', ``actually, let me'', ``hold on''.

\item \textbf{Planning}: ``we need'', ``first, i'', ``step 1'', ``step 2'', ``my approach'', ``i'll start by'', ``the plan is'', ``i need to'', ``i should'', ``let me start'', ``to solve this'', ``the strategy'', ``i will'', ``my strategy''.

\item \textbf{Computation}: ``calculating'', ``computing'', ``plugging in'', ``substituting'', ``evaluating'', ``simplifying'', ``expanding'', ``factoring'', ``multiplying'', ``dividing'', ``adding'', ``subtracting'', ``integrating'', ``differentiating'', ``solving for''.

\item \textbf{Fact retrieval}: ``we know that'', ``recall that'', ``remember that'', ``by definition'', ``by the formula'', ``using the formula'', ``the formula for'', ``it is known'', ``a known result'', ``from the theorem'', ``by theorem'', ``since we know''.
  \end{itemize}

\section{Sufficiency and attainability thresholds}
\label{app:suff-atnb-threshold-ablation}
We discuss our choice of sufficiency and attainability thresholds, and present supporting ablations.
In the main paper, both metrics share a notation $\tau = 0.8$; for clarity we give a separate notation ($\tau_{\text{suff}}, \tau_{\text{atnb}}$) for this appendix.

\paragraph{Attainability threshold.} 
Attainability is a continuous measure that controls how far the core reasoning subset may deviate from the model's distribution when reasoning steps are removed. 
A higher threshold restricts removal more strictly, keeping the remaining chain closer to in-distribution outputs.
Thus, to compare different values of $\tau_{\text{atnb}}$, we measure how the choice of threshold affects the prefix length $|\mathcal{T}_p|$ and the amount of removable and non-removable steps, as described in \Cref{sec:method}.

We ablate over $\tau_{\text{atnb}} \in \{0.6, 0.8, 0.9, 0.99\}$. 
Values below 0.6 lead to degenerate outputs, making attainability ineffective, so we ignore them.
Due to computational cost, we perform this analysis only on the Math-500 dataset.

\begin{figure}[h]
    \centering
    \includegraphics[width=\linewidth]{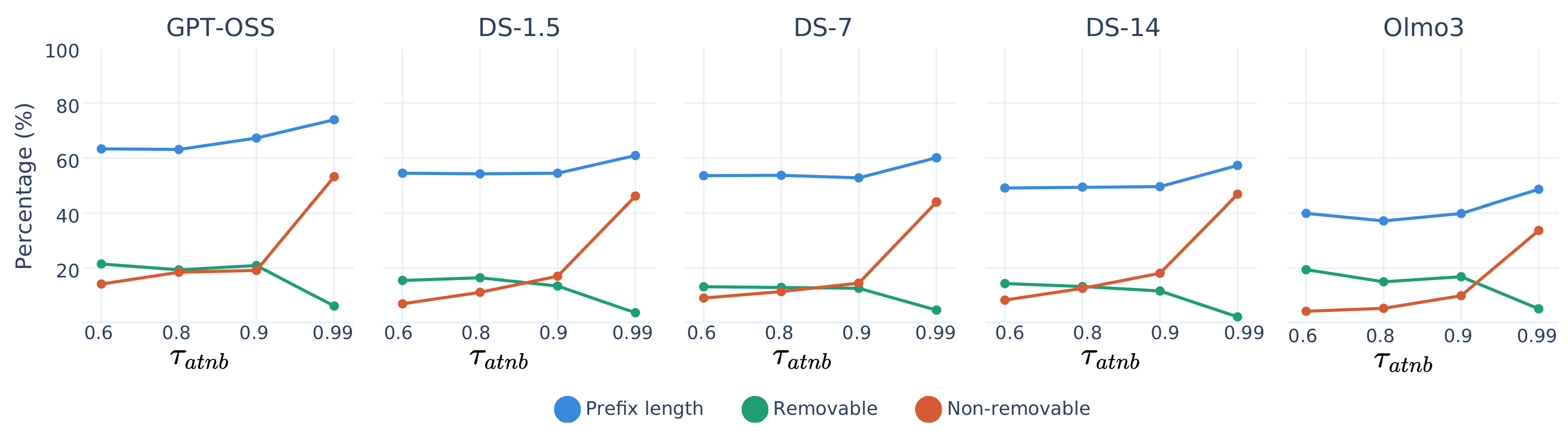}
    \caption{\textbf{The effect of different attainability thresholds ($\tau_{\text{atnb}}$) on the resulting lengths of removal experiments.}}
    \label{fig:attainability-threshold-length-ablations}
\end{figure}

For each threshold, we recompute the sufficient prefixes and rerun the activation-based variant (\Cref{sec:method}) on all prompts.
For each prompt, we record three quantities, each expressed as a percentage of the full reasoning chain: the prefix length in steps, the fraction of steps labeled removable, and the fraction labeled non-removable (i.e., the core reasoning subsequence size).
We average each quantity across prompts (per threshold and model) and report the results in \Cref{fig:attainability-threshold-length-ablations}

The results are consistent across models.
The prefix length (blue) stays nearly constant across thresholds, rising only at $\tau_{\text{atnb}} = 0.99$.
The core reasoning subset size (orange, non-removable steps) follows the same pattern. 
This is expected, as a stricter threshold limits how many steps the method can remove (green) before the chain goes out of distribution, which inflates the core subset with steps that are not strictly essential.

Due to the similar performance of the three analyzed thresholds $\tau_{\text{atnb}} \in \{0.6, 0.8, 0.9\}$, we default to choosing the middle $\tau_{atnb}=0.8$.
We opt to not use the more strict $\tau_{atnb}=0.99$ as it leads larger core reasoning subsets, that contain non-essential steps, which remain only to better match the model's output distribution.
We do note, however, that this is a tradeoff: applications that require a more in-distribution core reasoning chain could use the higher $\tau_{\text{atnb}}=0.99$ threshold.

\paragraph{Sufficiency threshold.} 
Sufficiency is measured as the fraction of correct answer out of $K=5$ completions. 
Thus, the set of relevant thresholds is $\tau_{\text{suff}} \in \{0.2, 0.4, 0.6, 0.8, 1.0\}$, each corresponding to $i \in [1,K]$ correct completions.
To focus on reasoning chains where the model answers correctly with high confidence, we only consider $\tau_{\text{suff}}\in \{0.8, 1.0\}$.
We choose $\tau_{\text{suff}} = 0.8$ as the sufficiency threshold for our analysis, corresponding to $i = 4$ correct answer out of $5$ completions, to tolerate occasional sampling errors while still requiring strong overall correctness.

\section{Performance statistics}
\label{app:performance-stats}

We report, for each model-dataset pair, the number of prompts that pass successive filtering stages in our analysis.
Specifically, we report five prompt counts per setting:
(i) the total number of analyzed prompts;
(ii) the number of prompts for which the model completed the generation of the reasoning chain (i.e. generation did not terminate due to token limit);
(iii) the number of prompts for which the reasoning chain is necessary;
(iv) the number of prompts for which the model produced a correct answer; and 
(v) the number of prompts for which the reasoning chain remains sufficient to obtain correct answer even after the removal of shortcut steps.
The full results are reported in \Cref{tab:prompt-stats}.

\begin{table}[h]
\centering
\begin{tabular}{llccccc}
\toprule
 &  & GPT-OSS & DS-1.5 & DS-7 & DS-14 & Olmo-3 \\
\midrule
\multirow{5}{*}{\textbf{HARP}} & Total & 5280 & 4886 & 4886 & 5280 & 4886 \\
 & Reasoning Complete & 4726 & 2752 & 3825 & 4019 & 3194 \\
 & Reasoning Necessary & 4360 & 2465 & 3219 & 3391 & 2838 \\
 & Correct Answer & 3578 & 2169 & 2808 & 3014 & 2539 \\
 & Shortcut-removed Suff. & 2430 & 726 & 938 & 1022 & 1087 \\
\midrule
\multirow{5}{*}{\textbf{Math-500}} & Total & 500 & 500 & 500 & 500 & 500 \\
 & Reasoning Complete & 479 & 370 & 437 & 453 & 408 \\
 & Reasoning Necessary & 397 & 295 & 314 & 339 & 324 \\
 & Correct Answer & 330 & 262 & 279 & 291 & 299 \\
 & Shortcut-removed Suff. & 235 & 114 & 130 & 121 & 156 \\
\bottomrule
\end{tabular}
\caption{\textbf{Number of prompts after each filtering, across models and datasets.}}
\label{tab:prompt-stats}
\end{table}

We also report, for each model-dataset pair, several step-level statistics in \Cref{tab:step-stats-harp,tab:step-stats-math}.
We report several metrics: 
(i) average number of steps in each reasoning chain; 
(ii) the average number of steps in the sufficient prefix, as a percentage of the total length; 
(iii-iv) the average number of steps labeled as removable and non-removable, as a percentage of the sufficient prefix length.

\begin{table}[h!]
\centering
\small
\begin{tabular}{lccccc}
\toprule
  & \textbf{GPT-OSS} & \textbf{DS-1.5} & \textbf{DS-7} & \textbf{DS-14} & \textbf{Olmo3} \\
\midrule
Avg.\ Steps & $96.5 \pm 75.8$ & $95.7 \pm 53.1$ & $99.4 \pm 57.3$ & $109.3 \pm 68.3$ & $191.4 \pm 90.2$ \\
Prefix Length & $64.2\% \pm 21.8$ & $56.1\% \pm 21.1$ & $51.2\% \pm 22.3$ & $55.2\% \pm 23.0$ & $34.1\% \pm 18.0$ \\
Removable & $37.6\% \pm 26.8$ & $27.3\% \pm 21.0$ & $28.1\% \pm 21.9$ & $30.6\% \pm 23.1$ & $36.4\% \pm 25.6$ \\
Non-removable & $22.3\% \pm 15.7$ & $24.4\% \pm 16.2$ & $23.4\% \pm 18.0$ & $25.4\% \pm 18.3$ & $20.1\% \pm 13.8$ \\
\bottomrule
\end{tabular}
\caption{\textbf{Reasoning step statistics (HARP dataset).} 
Prefix length is shown as percentage of total steps. Removable and non-removable are shown as percentage of prefix steps.}
\label{tab:step-stats-harp}
\end{table}

\begin{table}[h!]
\centering
\small
\begin{tabular}{llccccc}
\toprule
& \textbf{GPT-OSS} & \textbf{DS-1.5} & \textbf{DS-7} & \textbf{DS-14} & \textbf{Olmo3} \\
\midrule
Avg.\ Steps & $68.2 \pm 66.6$ & $79.8 \pm 46.4$ & $86.7 \pm 52.3$ & $87.3 \pm 57.9$ & $170.4 \pm 83.2$ \\
Prefix Length & $63.1\% \pm 22.2$ & $54.2\% \pm 22.6$ & $53.7\% \pm 21.0$ & $49.4\% \pm 21.0$ & $37.2\% \pm 19.0$ \\
Removable & $29.6\% \pm 24.1$ & $30.2\% \pm 21.5$ & $24.0\% \pm 21.2$ & $26.7\% \pm 22.6$ & $38.5\% \pm 26.0$ \\
Non-removable & $30.3\% \pm 16.9$ & $22.1\% \pm 13.0$ & $23.3\% \pm 17.3$ & $27.3\% \pm 15.0$ & $17.4\% \pm 11.3$ \\
\bottomrule
\end{tabular}
\caption{\textbf{Reasoning step statistics (MATH-500 dataset).}
Prefix length is shown as percentage of total steps. Removable and non-removable are shown as percentage of prefix steps.}
\label{tab:step-stats-math}
\end{table}

\section{Additional results}
\label{app:additional-results}

\subsection{Different input attribution methods}
\label{app:attribution-comparison}
In the activation-based variant (\Cref{sec:method}) we employ simple gradient attribution to measure influence between reasoning steps.
To ablate the effect of this choice on the importance labels assigned to reasoning steps, we compare it against alternative activation-based attribution methods.
Specifically, we compute $\mathbf{I}(s_i,s_j)$ using either gradient-input attribution \citep{shrikumar2017learning} or integrated gradients \citep{sundararajan2017axiomatic} with $n=5$ accumulation steps. The baseline used for integrated gradients is the average of all tokens in the reasoning chain.  
We evaluate each method by measuring the size of the core reasoning subsequence as a percentage of the sufficient prefix (identically to \Cref{tab:pruning-comparison}), and average across prompts.

Results on the Math-500 dataset are reported in \Cref{tab:attribution-comparison}.
All three attribution methods yield similar results, each outperforming both the token-based variant and the randomized variant in \Cref{sec:method}. 
While the integrated gradients method yields marginally smaller core reasoning subsequences, its improvement is small and comes at a substantially higher computational cost. We thus adopt the similarly-performing gradient-only scores across our remaining experiments. 

\begin{table}[h]
\centering
\resizebox{\textwidth}{!}{
\begin{tabular}{lccccc}
\toprule
Method & GPT-OSS & DS-1.5 & DS-7 & DS-14 & Olmo3 \\
\midrule
Gradient & $30.8\%\pm13.8$ & $22.1\%\pm13.0$ & $23.1\%\pm17.2$ & $27.8\%\pm13.6$ & $18.0\%\pm11.2$ \\
Gradient $\times$ Input & $26.9\%\pm13.9$ & $22.2\%\pm13.2$ & $21.0\%\pm18.0$ & $29.2\%\pm14.8$ & $19.7\%\pm16.2$  \\
Integrated Gradients & $24.3\%\pm11.4$ & $20.5\%\pm11.1$ & $19.2\%\pm14.7$ & $26.1\%\pm14.8$ & $18.1\%\pm12.3$  \\
\bottomrule
\end{tabular}
}
\caption{\textbf{Comparing the size of the resulting core reasoning subset using different gradient attribution methods.}}
\label{tab:attribution-comparison}
\end{table}

\subsection{Per-prompt comparison of core subsequence identification}
\label{app:more-per-prompt-variant-comparison}
\begin{figure}[h!]
    \begin{subfigure}{\linewidth}
        \includegraphics[width=\linewidth]{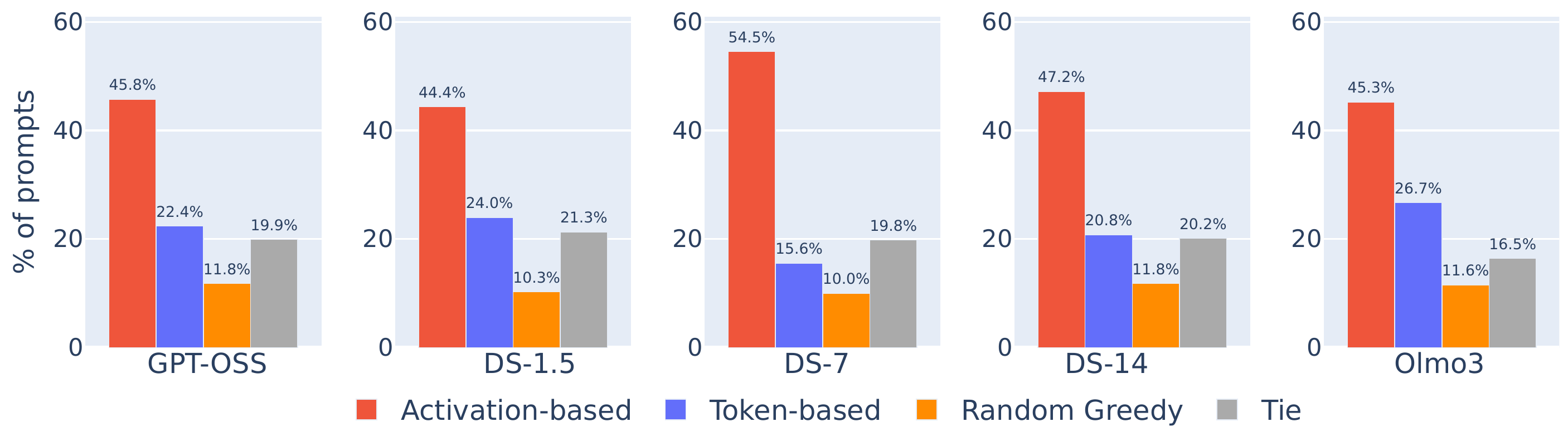}
        \caption{\textbf{Per-prompt comparison of the three variants on HARP.} We report the percentage of prompts for which each variant identifies the most concise core reasoning subset.}
        \label{fig:harp-per-prompt-variant-comparison}
    \end{subfigure}
    \vspace{4pt}
    \begin{subfigure}{\linewidth}
        \includegraphics[width=\linewidth]{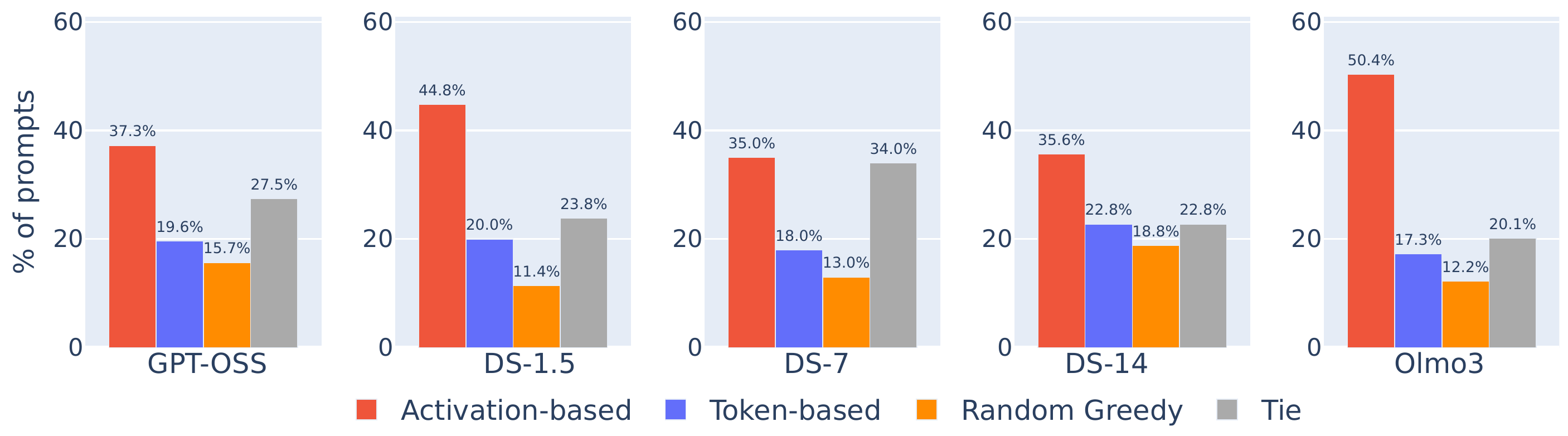}
        \caption{\textbf{Per-prompt comparison of the three variants on MATH-500.} We report the percentage of prompts for which each variant identifies the most concise core reasoning subset.}
        \label{fig:math-500-per-prompt-variant-comparison}
    \end{subfigure}
    \caption{}
    \label{fig:prompt-variant-comparison}
\end{figure}

In addition to reporting the average size of core reasoning subsequences $\mathcal{T}^*$ identified by each variant in \Cref{tab:pruning-comparison}, we compare the three variants per-prompt.
For each prompt, we check which variant yields the most $\mathcal{T}^*$.
We report the percentage of prompts for which each variant wins, along with the percentage of ties (\Cref{fig:prompt-variant-comparison}).
Across models and datasets, the activation-based variant, which utilizes gradient influence as an internal signal, yields the best results on the largest share of prompts. 
This is further motivation for our choice to use it to identify core reasoning subsets.

\newpage

\subsection{Activation probe accuracy under random-greedy labeling}
\label{app:probe-performance-random-labeling}
The activation-based probes presented in \Cref{sec:identifying-importance} are trained to predict importance labels derived from the causal process described in \Cref{sec:method}. These labels depend on two components: (i) the causal effect of removing a reasoning step, and (ii) the gradient-based ranking used to select removal candidates.
To isolate which of the two yields the signal learned by the probes, and to rule out potential circularity (arising from predicting gradient-based labels using activations), we train and evaluate probes in two settings that ablate the gradient-based ranking.

First, as a null control, we train probes on randomly assigned labels.
Second, we train probes on labels induced by the random-greedy pruning baseline (\Cref{sec:method}), in which removal candidates are selected randomly, rather than by gradient attribution.
Since the core reasoning subsequences resulting from the random-greedy baseline are comparable in length to those obtained by our gradient-based procedure (\Cref{tab:pruning-comparison}), these labels should provide a meaningful comparison.
In both ablations we train probes using the same setup as described in \Cref{sec:identifying-importance}.

We find that probes trained on fully random labels perform at chance level ($50\%$) across models and datasets, as expected.
In contrast, probes trained on random-greedy labels achieve accuracies comparable to the probes obtained using our original labels (\Cref{tab:random-labels-probe-results}). 
These results indicate that the signal captured by the probes does not originate from the gradient-based ranking, but rather the underlying causal property: whether a step can be removed while maintaining high sufficiency and attainability.

\begin{table}[h!]
\centering
\begin{tabular}{lccccc}
\toprule
Dataset & GPT-OSS & DS-1.5B & DS-7B & DS-14B & Olmo3 \\
\midrule
HARP     & 82.6\% & 82.1\% & 81.4\% & 79.6\% & 80.9\% \\
MATH-500 & 80.1\% & 78.1\% & 77.8\% & 74.9\% & 77.1\% \\
\bottomrule
\end{tabular}
\caption{Probe accuracy when trained on labels from the random-greedy baseline. 
The resulting accuracies are comparable to those obtained with out original labels.}
\label{tab:random-labels-probe-results}
\end{table}

\subsection{Reasoning step examples and labels}
\label{app:reasoning-step-instances}

We present qualitative examples of importance labels for individual reasoning steps.
These examples are organized into $4$ groups, defined by the combination of their ground-truth label (removable or non-removable) and activation probe prediction success (\Cref{sec:activation-importance-exp}).
Each group is colored as light red if the steps within it are removable, or light green if they are important and non-removable. 
The title of each group indicates if the full-context activation probes correctly classified the group's steps.  

The steps are \emph{randomly sampled} from the GPT-OSS-20B analysis on MATH-500.
Full results, including all labeled steps in all experimental settings, are available in the accompanying data repository.

These examples illustrate that surface-level features alone are often insufficient for predicting importance: steps of the same surface type---fillers, direct computations, and others---can appear with different ground truth importance labels.

\begin{mdframed}[backgroundcolor=red!10, nobreak=true, frametitle={GT: Removable | Activation Probe Correct: $\checkmark$}]
\begin{enumerate}
  \item Actually parallelogram sides: one side is vertical segment between (0,c) and (0,d).
  \item Let's analyze.
  \item Let's analyze region.
  \item So sign of A and B?
  \item We can write as $e^{i\theta}(e^{-i\theta} -1 + e^{i\theta}) = e^{i\theta}(1 - e^{i\theta} + e^{i2\theta})$?
  \item Term1: $(x - z)^2 / 2 = (x^2 - 2xz + z^2)/2$. Good.
  \item But we must check: Did we compute correctly?
  \item Solve congruences.
  \item So AC direction vector ($x_C$, h) is bisector of angle between BA and AD.
  \item Wait sign: -4xy?
  \item 13/9 = 52/36.
  \item Similarly odd: $\Pi_{k=1}^n ((2k-1)^2 + (2k-1) + 1/2)((2k-1)^2 - (2k-1) + 1/2)$.
  \item So we have factorization: $x^4 + 1/4 = (x^2 + x + 1/2)(x^2 - x + 1/2)$. Good.
  \item So include.
  \item So we need to find intersection with AC: G + t(c-b) = u c.
\end{enumerate}
\end{mdframed}

\begin{mdframed}[backgroundcolor=red!10, nobreak=true, frametitle={GT: Removable | Activation Probe Correct: $\times$}]
\begin{enumerate}
  \item So x must be $\ge (3+sqrt33)/2 \approx 4.372$.
  \item Because sum of 0 to 26 = 26*27/2 = 351.
  \item Without loss of generality, we can choose F to be a particular face, say 1?
  \item We can solve.
  \item But we should also check interior points maybe produce lower?
  \item But we need actual integer values.
  \item $r^2$ must be one of these.
  \item So square: $4(3x+6) = 4(3x+6)$. trivial.
  \item So compare coefficients:
  \item So $\approx0.4097 <1$. Good.
  \item Then compute their average.
  \item a/(b+c) = a/(c(v+1)) = (a/c)/(v+1) = u/(v+1).
  \item But f is not one-to-one globally.
  \item Area = $\int_{x=0}^{1} [ (x+1) - (1 - x) ] dx + \int_{x=1}^{2} [ (x+1) - (x-1) ] dx + \int_{x=2}^{3} [ 3 - (x-1) ] dx.$
  \item Then $u^2 = v^2 -9 = 36 \pm 4 sqrt41$.
\end{enumerate}
\end{mdframed}

\begin{mdframed}[backgroundcolor=green!10, nobreak=true, frametitle={GT: Non-removable | Activation Probe Correct: $\checkmark$}]
\begin{enumerate}
  \item So equation: $6r^2 - 19r - 7 = 8r^2 - 34r + 21$.
  \item So compute: $(3 - (-4-5i)) = 3 + 4 + 5i = 7 + 5i$.
  \item Thus area of triangle $ACD = (1/2)*AD*CD = (1/2)*(64/17)*(120/17) = (1/2)*(64*120)/(17^2)$.
  \item $555 < 625$, so highest power is $5^3=125$.
  \item So $z1 = (a + 2i)/a = 1 + (2i)/a$.
  \item We need to maximize $f(x) = 4(x+7)(2-x)$.
  \item - above line x + y = 1 - above line y = x - 1 - below line y = x + 1 - within square.
  \item Shoelace $sum_1 = x_i*y_{i+1}: -1*4 = -4 2*(-4) = -8 2*(-1) = -2 -1*1 = -1 Sum_1 = -4-8-2-1 = -15$
  \item Let S = 3n+3 = 3(n+1).
  \item Compute 5x+9 at x=-2: 5*(-2)+9 = -10+9 = -1.
  \item Let's compute vectors.
  \item For i=2, j=2: f(2,2)=f(1, f(2,1))=f(1,0)=2. Yes.
  \item So remainder 3.
  \item For x between 0 and 1, y between 1 and 0.
  \item We need count of pairs (a,b) such that a*b divisible by 5.
\end{enumerate}
\end{mdframed}

\begin{mdframed}[backgroundcolor=green!10, nobreak=true, frametitle={GT: Non-removable | Activation Probe Correct: $\times$}]
\begin{enumerate}
  \item not perfect square.
  \item So quotient $x^2 -4x -12$.
  \item Check if within interval [-1,1/2]. Yes.
  \item Compute $a = sqrt(20^2+400)= sqrt(400+400)= sqrt(800)= 20*sqrt(2)=28.2843$.
  \item That gives u = 39, 89, 139, etc.
  \item - For $i\ge1:   - if j=0: f(i,0) = f(i-1,1)$.
  \item $G - M = ((b + c_x)/3 - 2b/3, c_y/3 - 0) = ((b + c_x - 2b)/3, c_y/3) = ((c_x - b)/3, c_y/3)$.
  \item Wait $8^2+15^2=64+225=289=17^2$. Yes!
  \item Equation same: $kx = (x+2)^2$.
  \item So $4x= 16y -25$.
  \item The problem says: The ratio of Denali's pay to Nate's pay would be the same if Denali started walking 4x more dogs and Nate stayed at 12 dogs or if x of Nate's dogs were reassigned to Denali.
  \item Now Archimedes wants to minimize f(x).
  \item sqrt(3481) =?
  \item Next $8^{2/3}$.
  \item Since 1985 is odd, we can compute: R = 1 - 2 + 3 - 4 + ... + 1985.
\end{enumerate}
\end{mdframed}

\subsection{Reasoning step importance prediction results}
\label{app:more-results-for-classification}

\begin{figure}[h!]
    \centering
    \includegraphics[width=\linewidth]{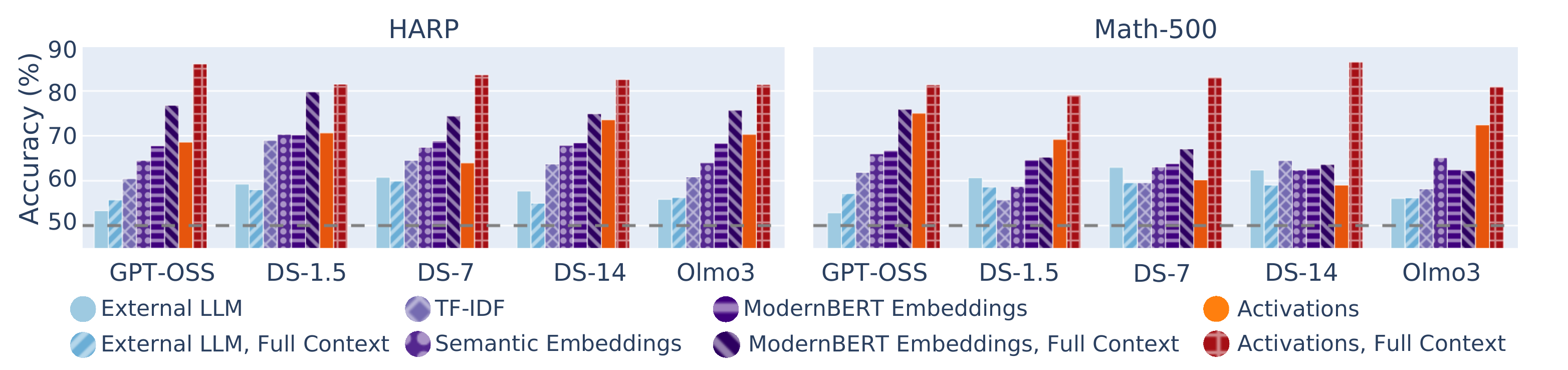}
    \caption{\textbf{Evaluation of all importance detection methods on the $1500$-sized LLM-as-a-judge evaluation subset.}}
    \label{fig:more-results-for-classification}
\end{figure}

In \Cref{sec:identifying-importance}, we define and evaluate several methods for predicting reasoning step importance from tokens and latent representations.
Due to budget constraints on the number of steps that could be processed through the external LLM-as-a-judge, we analyze this baseline on a limited evaluation set of randomly-chosen $1,500$ steps per model and dataset.
The results of non-LLM-as-a-judge methods, presented in \Cref{fig:importance-classification}, report evaluation accuracy on all available data.
For an identical comparison, we show in \Cref{fig:more-results-for-classification} the results of the other methods when evaluated only on the $1,500$ steps.
The results are consistent with those in \Cref{sec:identifying-importance} across all models and datasets.

\subsection{Math-500 cross-model results}
\label{app:math-crossmodel-additional-results}
\begin{figure}[h!]
    \centering
    \includegraphics[width=\linewidth]{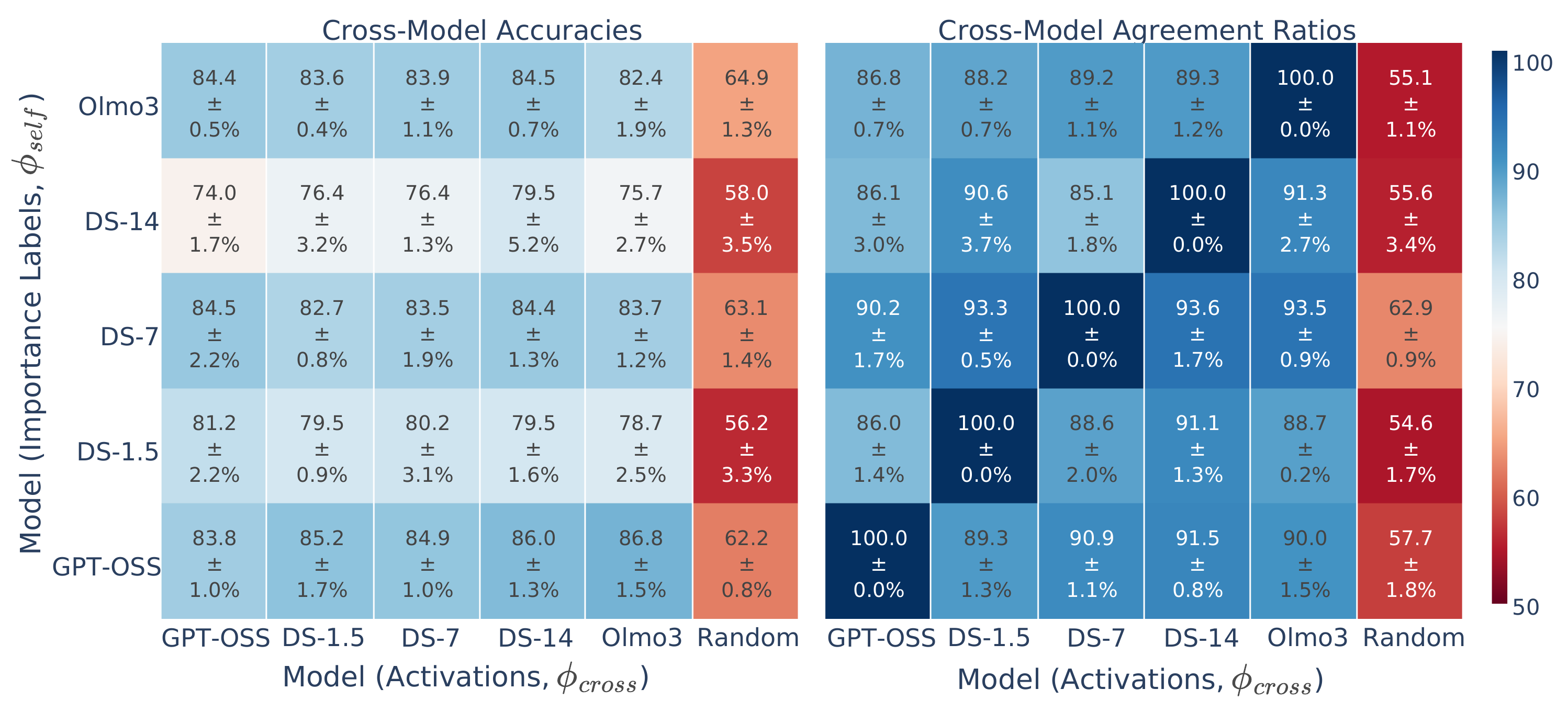}
    \caption{\textbf{Cross-model activation-based probe results on MATH-500.}
    Probes trained on one model's activations ($\phi_\text{cross}$) are able to predict labels derived from another model ($\phi_\text{self}$), and have high agreement ratios on individual labels with the probe trained when $\phi_\text{cross}=\phi_\text{self}$, suggesting the notion of importance is encoded in a universal manner across mathematical reasoning datasets.
    }
    \label{fig:math-500-cross-model-results}
\end{figure}

We show results for the cross-model analysis on the MATH-500 dataset in \Cref{fig:math-500-cross-model-results}.
The results for MATH-500 are consistent with the results presented in \Cref{fig:cross-model-results}, supporting the claim that the notion of importance is universal across models.   

\subsection{Cross-model importance representation alignment}
\label{app:cross-model-importance-alignment}

In \Cref{sec:cross-model-importance} we showed that probes trained on activations from one model can predict importance labels from another, suggesting different models agree on which reasoning steps are important, through their activations.
This raises a related question: is importance encoded similarly in the activation of different models?
To answer this, we test whether a probe trained on activations from one model $\phi_{tgt}$ can correctly predict importance from activations from another model $\phi_{src}$.%

Because different models differ in hidden dimensions and layer counts, and their representations are likely non-identical, we learn an alignment between activation spaces.
For each model pair $\phi_{src},\phi_{tgt}$ we train a linear mapping $M_{src \rightarrow tgt} \in \mathbb{R}^{d_{tgt} \times d_{src}}$ between their penultimate-layer hidden states. We use the penultimate layer to avoid mismatched layer counts across models. A probe trained only on this layer achieves accuracy close to our all-layer probe (\Cref{app:more-probe-analysis-results}).

To train $M_{src \rightarrow tgt}$ we collect activations for each reasoning step generated by $\phi_{src}$.
For each step, we feed its full prompt to both models, extract the step's activations at the penultimate layer, and average them across tokens.
The mapping is trained with an MSE loss, and evaluated on a held-out evaluation set of reasoning step activations. 
We validate the mapping reaches high mean cosine similarities ($\ge0.98$) on the evaluation set.
Finally, use map activations of $\phi_{src}$ and evaluate the probe trained on $\phi_{tgt}$ activations, to measure cross-model representation alignment.

The results (\Cref{tab:cross-model-aligned-probe-transfer}) show lower accuracies when evaluating a probe on another model's aligned activations. Notably, when both $\phi_{src},\phi_{tgt}$ are within the DeepSeek family, transfer accuracy is higher compared to transfer accuracy across model families.
This may point towards different captured representations of importance across models, despite these different representations pointing towards similar labels (\Cref{fig:cross-model-results}).
However, an alternative explanation is that the linear alignment, despite producing high cosine similarity, does not fully recover the subspaces used by the activation-based probes.

\begin{table}[h!]
\centering
\small
\begin{tabular}{llccccc}
\toprule
& & \multicolumn{5}{c}{$\phi_{\mathrm{tgt}}$} \\
\cmidrule(lr){3-7}
Dataset & $\phi_{src}$ & GPT-OSS & DS-1.5B & DS-7B & DS-14B & Olmo3 \\
\midrule
\multirow{5}{*}{HARP}
 & GPT-OSS & 78.3\% & 58.5\% & 61.6\% & 59.6\% & 53.9\% \\
 & DS-1.5B & 50.1\% & 78.2\% & 68.7\% & 68.7\% & 63.0\% \\
 & DS-7B   & 53.0\% & 67.8\% & 79.3\% & 68.6\% & 62.5\% \\
 & DS-14B  & 55.2\% & 68.8\% & 68.3\% & 81.3\% & 60.4\% \\
 & Olmo3   & 50.3\% & 64.1\% & 66.4\% & 66.5\% & 76.6\% \\
\midrule
\multirow{5}{*}{MATH-500}
 & GPT-OSS & 78.1\% & 57.3\% & 58.1\% & 52.8\% & 53.1\% \\
 & DS-1.5B & 50.8\% & 80.2\% & 69.9\% & 65.3\% & 63.1\% \\
 & DS-7B   & 51.0\% & 68.1\% & 82.5\% & 65.7\% & 65.0\% \\
 & DS-14B  & 52.6\% & 67.6\% & 67.5\% & 79.3\% & 60.2\% \\
 & Olmo3   & 50.1\% & 60.1\% & 63.3\% & 62.3\% & 78.3\% \\
\bottomrule
\end{tabular}
\caption{Probes trained on activations and labels from $\phi_{tgt}$ yield lower accuracies when evaluated on aligned activations from $\phi_{src}$, indicating that importance is encoded differently in different models.}
\label{tab:cross-model-aligned-probe-transfer}
\end{table}

\subsection{Probe analysis results}
\label{app:more-probe-analysis-results}

\begin{figure}[h!]
    \centering
    \includegraphics[width=.95\linewidth]{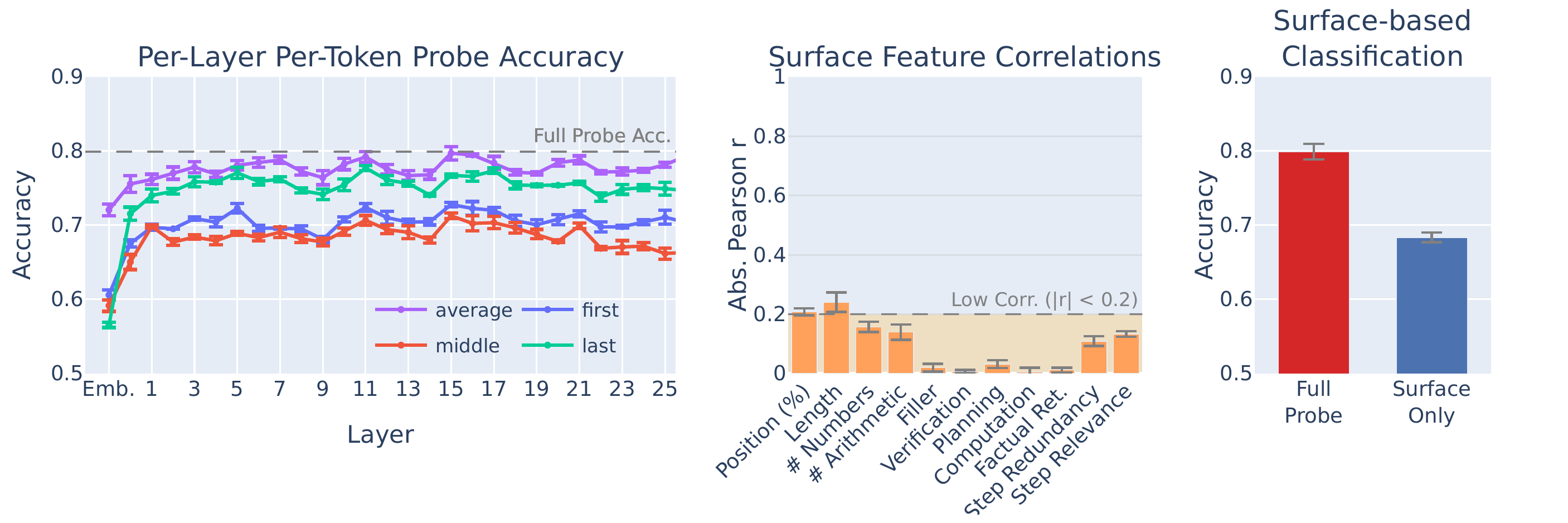}
    \caption{\textbf{Analysis of the importance probe in DeepSeek-1.5B (HARP dataset).}}
    \label{fig:deepseek-15b-probe-analysis}
\end{figure}

\begin{figure}[h!]
    \centering
    \includegraphics[width=.95\linewidth]{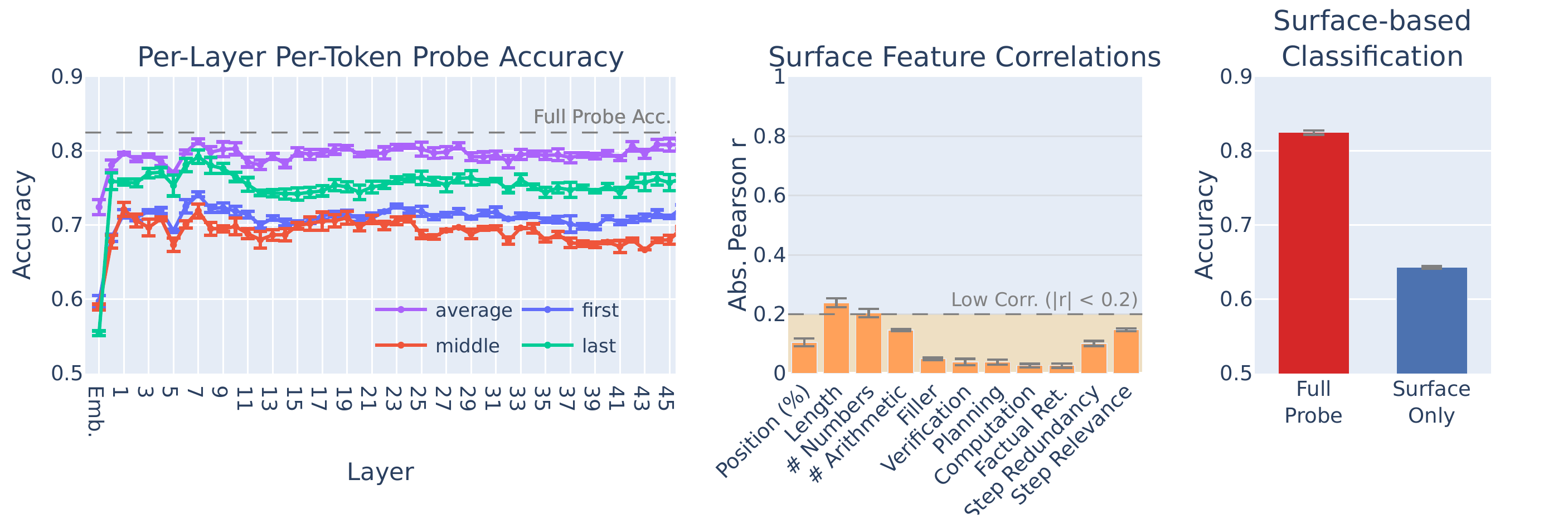}
    \caption{\textbf{Analysis of the importance probe in DeepSeek-14B (HARP dataset).}}
    \label{fig:deepseek-14b-probe-analysis}
\end{figure}

\begin{figure}[h!]
    \centering
    \includegraphics[width=.95\linewidth]{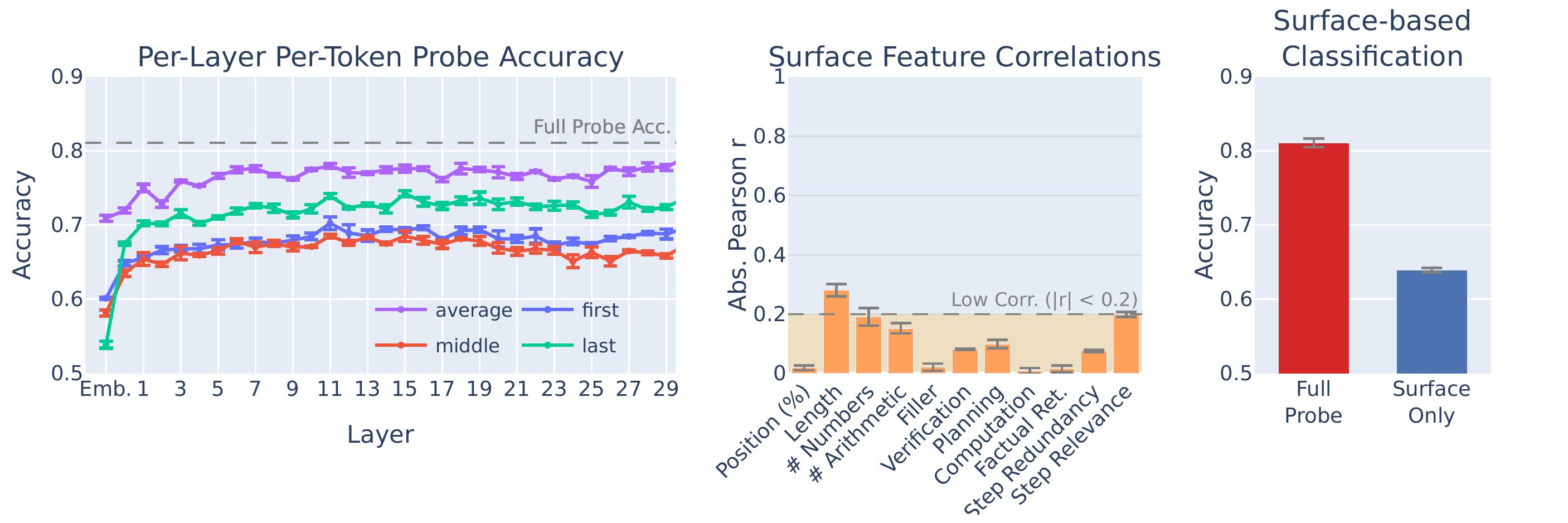}
    \caption{\textbf{Analysis of the importance probe in Olmo3-7B-Think (HARP dataset).}}
    \label{fig:olmo-3-probe-analysis}
\end{figure}

\begin{figure}[h!]
    \centering
    \includegraphics[width=.95\linewidth]{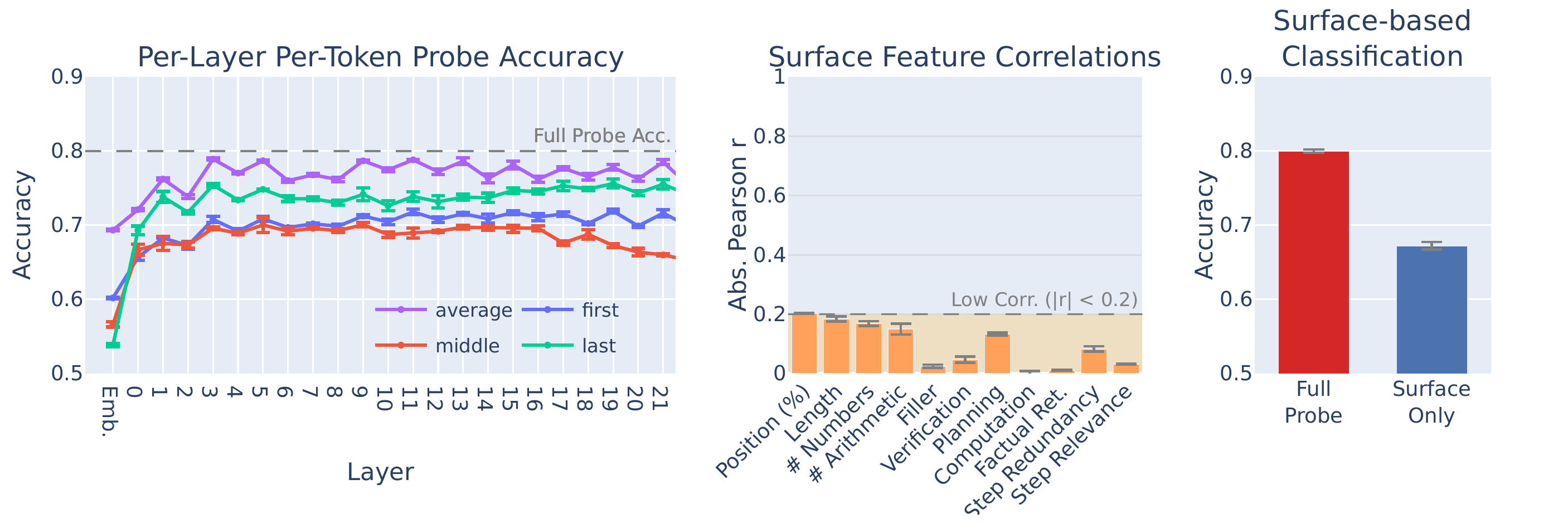}
    \caption{\textbf{Analysis of the importance probe in GPT-OSS-20B (HARP dataset).}}
    \label{fig:gpt-oss-20b-probe-analysis}
\end{figure}

\Cref{fig:deepseek-15b-probe-analysis,fig:deepseek-14b-probe-analysis,fig:olmo-3-probe-analysis,fig:gpt-oss-20b-probe-analysis} present the probe analysis results on all models, as discussed in \Cref{sec:concept-analysis}.
Each figure contains three sub-plots.
The left sub-plot shows accuracies of alternative probes---trained on activations from different token positions within each reasoning step, at different layers.
The central sub-plot reports the correlation of probe predictions with surface-level features, which is consistently low across all models and probes. 
The right sub-plot compares probe accuracy against a regression model trained on all surface-level features: activation-based probes outperform this baseline by a large margin, further supporting the conclusion that the probes capture meaningful, non-surface-level structure.

\subsection{Negative results}
\label{app:negative-results}

Several of our experiments yielded negative results that did not contribute to our understanding of the concept of reasoning step importance, yet we view them as worthwhile to mention.

\subsubsection{Different intervention types}
To investigate whether specific reasoning steps have different functional roles in the reasoning process---and whether these relate to their importance---we design a set of counterfactual interventions applied to individual steps.
Unlike the sufficiency evaluations in the main paper, where all steps except a candidate subsequence are removed, here we intervene on a \emph{single} step at a time while leaving the rest of the chain intact.

For each target step, we apply four interventions:
\begin{enumerate}
\item \textbf{Removal}: the step is deleted entirely from the reasoning chain. This is the step-level parallel to the removal intervention from the main paper.

\item \textbf{Filler}: the step is replaced with repetitions of the filler sentence ``Let me think...'' padded with \texttt{...} tokens to match the original token count.

\item \textbf{Semantic paraphrase}: the step is replaced with a semantically equivalent restatement generated by an external LLM (Gemini-2.5-Flash), preserving meaning while varying its tokens.

\item \textbf{Context-averaged activation}: the step's activations are replaced with its context-averaged activations. 
To calculate the context-averaged activations, we append the target reasoning step to $10$ randomly sampled reasoning chains, after a randomly chosen reasoning step. We calculate the target reasoning step's activations and average them across those contexts.
\end{enumerate}

Across all model–dataset settings, the four interventions produce highly correlated outcomes.
In a vast majority of cases, no single-step intervention has a meaningful effect on sufficiency.
We attribute this to the inherent redundancy of reasoning chains: even if an important step is intervened on, the model can still recover the correct answer from other steps.
These results further motivate our choice to intervene on reasoning chains in the reasoning subsequence resolution.

To measure a more fine-grained effect of these interventions, we measure the $l_2$ distance between the pre- and post-intervention activation tensors of reasoning steps subsequent to the intervened step.
All four interventions again show consistent correlations with one another: when one intervention produces a large downstream perturbation, so do the others, this difference does not localize to specific step types in a way that supports a meaningful taxonomy.
Removal and filler interventions generally produce larger perturbations than the paraphrase or context-averaged interventions, but these differences do not localize to specific steps in a way that supports a meaningful taxonomy of step types based on functionality.

\newpage

\subsubsection{Analysis of importance probes} 

To assess whether the probe weights encoded interpretable directions in the residual stream, we applied the logit lens \citep{nostalgebraist2020logit} to the per-layer weight vectors of the layerwise-linear probe (each of shape $\mathbb{R}^d$). 
Across all model–dataset pairs and layers, the projected tokens were not semantically meaningful, suggesting the probes do not operate through easily human-interpretable vocabulary directions (See results for a subset of layers in \Cref{tab:logit-lens}).

\begin{table}[ht]
\centering
\resizebox{\textwidth}{!}{
\small
\begin{tabular}{c l l}
\toprule
\textbf{Layer} & \textbf{Direction} & \textbf{Tokens} \\
\midrule
\multirow{2}{*}{0} & Top & \texttt{aea}, \texttt{agua}, \texttt{.cool}, \texttt{cko}, \texttt{cheon}, \texttt{elman}, \texttt{(lang}, \texttt{.random}, \texttt{.scroll}, \texttt{.arm} \\
 & Bottom & \texttt{arre}, \texttt{mout}, \texttt{Jules}, \texttt{spl}, \texttt{ozn}, \texttt{\_PUSH}, \texttt{blaze}, \texttt{Published}, \texttt{Written}, \texttt{Ment} \\
\addlinespace
\multirow{2}{*}{3} & Top & \texttt{qat}, \texttt{eness}, \texttt{/modal}, \texttt{inam}, \texttt{nae}, \texttt{Quin}, \texttt{quin}, \texttt{yna}, \texttt{quiv}, \texttt{enk} \\
 & Bottom & \texttt{arre}, \texttt{-Star}, \texttt{soak}, \texttt{URE}, \texttt{softer}, \texttt{tramp}, \texttt{Fle}, \texttt{cem}, \texttt{LEG}, \texttt{HG} \\
\addlinespace
\multirow{2}{*}{7} & Top & \texttt{.fm}, \texttt{akus}, \texttt{idad}, \texttt{kau}, \texttt{-Bo}, \texttt{ista}, \texttt{.Serv}, \texttt{aua}, \texttt{Pier}, \texttt{alur} \\
 & Bottom & \texttt{dise}, \texttt{goo}, \texttt{Invalidate}, \texttt{uret}, \texttt{Lap}, \texttt{Hond}, \texttt{ADVISED}, \texttt{vii}, \texttt{reneg}, \texttt{invalidate} \\
\addlinespace
\multirow{2}{*}{11} & Top & \texttt{accommodating}, \texttt{\_logic}, \texttt{circul}, \texttt{\_UPLOAD}, \texttt{\_plugins}, \texttt{imaginative}, \texttt{mou}, \texttt{mos}, \texttt{akter}, \texttt{konst} \\
 & Bottom & \texttt{ysa}, \texttt{asal}, \texttt{DF}, \texttt{SF}, \texttt{Nether}, \texttt{incer}, \texttt{ELE}, \texttt{chl}, \texttt{APS}, \texttt{IPP} \\
\addlinespace
\multirow{2}{*}{15} & Top & \texttt{PRES}, \texttt{-Str}, \texttt{-now}, \texttt{relate}, \texttt{pesc}, \texttt{UDIO}, \texttt{\_GENERIC}, \texttt{TEC}, \texttt{prescribe}, \texttt{ICES} \\
 & Bottom & \texttt{hein}, \texttt{vien}, \texttt{afr}, \texttt{etto}, \texttt{anoa}, \texttt{oron}, \texttt{\%+}, \texttt{aign}, \texttt{court}, \texttt{Him} \\
\addlinespace
\multirow{2}{*}{19} & Top & \texttt{MUT}, \texttt{alja}, \texttt{Haarlem}, \texttt{Hughes}, \texttt{Cartoon}, \texttt{heads}, \texttt{HQ}, \texttt{perfectly}, \texttt{accr}, \texttt{GBR} \\
 & Bottom & \texttt{markdown}, \texttt{tala}, \texttt{mete}, \texttt{lade}, \texttt{.pattern}, \texttt{avana}, \texttt{pomp}, \texttt{ond}, \texttt{meteor}, \texttt{maq} \\
\addlinespace
\multirow{2}{*}{23} & Top & \texttt{037}, \texttt{770}, \texttt{944}, \texttt{071}, \texttt{tay}, \texttt{cind}, \texttt{210}, \texttt{799}, \texttt{taan}, \texttt{pone} \\
 & Bottom & \texttt{Moc}, \texttt{[d}, \texttt{dj}, \texttt{(d}, \texttt{=d}, \texttt{d}, \texttt{Moj}, \texttt{hv}, \texttt{roller}, \texttt{Mog} \\
\bottomrule
\end{tabular}
}
\caption{Logit lens analysis of specific layers in layerwise-linear probe weights. Each layer's weight vector is projected through the model's unembedding to identify the most positively (Top) and negatively (Bottom) associated tokens.}
\label{tab:logit-lens}
\end{table}

\newpage
\section{Reasoning chain examples}
\label{app:reasoning-chain-examples}
We present several examples for generated reasoning chains.
For each reasoning chain, we mark shortcut steps that were removed pre-analysis in \textcolor{gray}{gray}, removable reasoning steps in \textcolor{orange}{orange} and non-removable reasoning steps in \textcolor{OliveGreen}{green}.
The minimal sufficient prefix, after which the model is able to generate the answer with high probability, is marked in \textbf{bold}.
For presentation purposes, the examples shown here are relatively short reasoning chains (less than $20$ reasoning steps and $1000$ tokens). 
The full results, including the generated reasoning chains and the importance analysis results are available in the accompanying data repository.

\begin{qabox}{GPT-OSS-20B, HARP}
    \textbf{Input:} Solve the following problem. You MUST put your final answer in \textbackslash boxed\{\}. \\
    Problem: The smallest product one could obtain by multiplying two numbers in the set $\{ -7,-5,-1,1,3 \}$ is:
  \tcblower          %
  \textbf{Reasoning Chain:} 
  \texttt{
  \textbf{
    \textcolor{OliveGreen}{We need to find smallest product.}
    \textcolor{orange}{most negative? or smallest numeric value? Usually "smallest product" meaning minimal value, i.e., most negative.
    We have set of numbers: -7, -5, -1, 1, 3.
    Multiply any two numbers (could be same? Usually two numbers from set, maybe distinct? Not specified. Usually "multiplying two numbers in the set" could be any two numbers, maybe same allowed? But product of same number with itself? 
    Let's consider both possibilities.
    We need smallest product.}
    \textcolor{OliveGreen}{Let's compute all pairwise products.
    List numbers: -7, -5, -1, 1, 3.
    Compute products:
    -7 * -5 = 35, -7 * -1 = 7, -7 * 1 = -7, -7 * 3 = -21, -5 * -1 = 5, -5 * 1 = -5, -5 * 3 = -15, -1 * 1 = -1, -1 * 3 = -3, 1 * 3 = 3.
    Also consider squares: -7 * -7 = 49, -5 * -5 = 25, -1 * -1 = 1, 1*1=1, 3*3=9.}
    }
    \textbf{\textcolor{orange}{Now find smallest value among all these: we have negative values: -21, -15, -7, -5, -3, -1.}}
    \textcolor{gray}{The smallest is -21.}
    \textcolor{gray}{So product -21 from -7 * 3}. 
    Also -7 * 3 = -21. 
    Also -5 * 3 = -15. 
    So minimal is -21. But check if any product less than -21? No. 
    \textcolor{gray}{So answer -21.} 
    But if we consider distinct numbers only, -7 * 3 is distinct, so still -21. 
    \textcolor{gray}{So answer -21.}
    \textcolor{gray}{Thus boxed answer: \textbackslash boxed\{-21\}}.
  }
\end{qabox}

\begin{qabox}{GPT-OSS-20B, HARP}
    \textbf{Input:} Solve the following problem. You MUST put your final answer in \textbackslash boxed\{\}. \\
    Problem: For $p=1, 2, \cdots, 10$ let $S_p$ be the sum of the first $40$ terms of the arithmetic progression whose first term is $p$ and whose common difference is $2p-1$; then $S_1+S_2+\cdots+S_{10}$ is:
  \tcblower          %
  \textbf{Reasoning Chain:} 
  \texttt{
  \textbf{We need to compute \boldmath{$S_p$} for each \boldmath{$p$} from 1 to 10.} 
  \textbf{\textcolor{orange}{Arithmetic progression: first term \boldmath{$a_1 = p$}, common difference \boldmath{$d = 2p-1$}.}} 
  \textbf{\textcolor{orange}{Number of terms \boldmath{$n = 40$}.}} 
    \textbf{\textcolor{OliveGreen}{Sum formula: \boldmath{$S_p = n/2 * (2a1 + (n-1)d) = 40/2 * (2p + (39)(2p-1))$}.}} 
    \textbf{\textcolor{OliveGreen}{Let's compute: \boldmath{$20 * (2p + 39(2p-1))$}.}} 
    \textbf{\textcolor{orange}{Simplify inside: \boldmath{$2p + 39(2p-1) = 2p + 78p - 39 = 80p - 39$}.}} 
    \textbf{\textcolor{OliveGreen}{So \boldmath{$S_p = 20 * (80p - 39) = 1600p - 780$}.}} 
    \textbf{\textcolor{orange}{Wait \boldmath{$20*80p = 1600p; 20*(-39) = -780$}.}} 
    \textbf{\textcolor{OliveGreen}{So \boldmath{$S_p = 1600p - 780$}.}} 
    \textbf{\textcolor{OliveGreen}{Now sum over \boldmath{$p=1$} to \boldmath{$10$}: Sum \boldmath{$S = \sum_\{p=1\}^\{10\} (1600p - 780) = 1600 \sum p - 780*10$}.}} 
    \textbf{\textcolor{orange}{\boldmath{$\sum p from 1 to 10 = 55$}.}} 
    \textbf{\textcolor{OliveGreen}{So \boldmath{$1600*55 = 1600*50 + 1600*5 = 80,000 + 8,000 = 88,000$}.}} 
    \textbf{\textcolor{OliveGreen}{Then minus \boldmath{$7800 = 80,200$}?}}
    Wait $780*10 = 7,800$. 
    \textcolor{gray}{So $88,000 - 7,800 = 80,200$.}
    \textcolor{gray}{So answer $80200$.}
    Let's double-check arithmetic: $1600*55 = 1600*(5*11)$??
    Actually $55 = 5*11$. $1600*5 = 8000$; $8000*11 = 88,000$. Yes. Subtract $7,800$ gives $80,200$. 
    \textcolor{gray}{So final answer: $80200$.}
    Let's ensure no misinterpretation: $S_p$ is sum of first $40$ terms of AP with first term $p$ and common difference $2p-1$.
    Yes. So $S_p = 20*(2p + 39(2p-1))$.
    Let's recompute inside: $2p + 39(2p-1) = 2p + 78p - 39 = 80p - 39$. Good. Multiply by $20$: $1600p - 780$. Good. 
    \textcolor{gray}{Sum $p=1$ to $10$: $1600*55 - 780*10 = 88,000 - 7,800 = 80,200$. Good.}
    So answer: $80200$.
    Provide in boxed form.
  }
\end{qabox}

\begin{qabox}{DeepSeek-R1-7B, HARP}
  \textbf{Input:} Solve the following problem. You MUST put your final answer in \textbackslash boxed\{\}. \\
    Problem: Chandler wants to buy a $500$ dollar mountain bike. For his birthday, his grandparents send him $50$ dollars, his aunt sends him $35$ dollars and his cousin gives him $15$ dollars. He earns $16$ dollars per week for his paper route. He will use all of his birthday money and all of the money he earns from his paper route. In how many weeks will he be able to buy the mountain bike?
  \tcblower          %
  \textbf{Reasoning Chain:} 
  \texttt{
    \textbf{\textcolor{OliveGreen}{Okay, so Chandler wants to buy a mountain bike that costs \$500.}} 
    \textbf{\textcolor{orange}{He gets some money from his grandparents, aunt, and cousin for his birthday, and he also earns some money from his paper route every week.}} 
    \textbf{\textcolor{orange}{The question is asking how many weeks he needs to work to earn enough money to buy the bike, using all his birthday money and all his weekly earnings.}} 
    \textbf{\textcolor{OliveGreen}{First, let me figure out how much birthday money Chandler has in total.}} 
    \textbf{\textcolor{OliveGreen}{His grandparents sent him \$50, his aunt sent \$35, and his cousin gave \$15.}} 
    \textbf{\textcolor{orange}{So, I need to add those amounts together.}} 
    \textbf{Let me write that down: 50 + 35 + 15.} 
    \textbf{\textcolor{OliveGreen}{Hmm, 50 plus 35 is 85, and then 85 plus 15 is 100.}} 
    \textbf{\textcolor{OliveGreen}{So, Chandler has \$100 from his birthday gifts.}} 
    \textbf{\textcolor{OliveGreen}{Now, the bike costs \$500, and he already has \$100.}} 
    \textbf{\textcolor{OliveGreen}{So, I need to find out how much more money he needs.}} 
    \textbf{\textcolor{orange}{That would be 500 minus 100, which is 400.}} 
    \textbf{\textcolor{OliveGreen}{So, Chandler still needs \$400.}} 
    \textbf{\textcolor{OliveGreen}{He earns \$16 per week from his paper route.}} 
    He's going to use all of his birthday money and all of his weekly earnings to buy the bike. \textcolor{gray}{So, the \$400 he needs will come entirely from his weekly earnings because he's already used up all his birthday money.} Now, to find out how many weeks he needs to work to earn \$400 at \$16 per week, I can set up an equation. Let me denote the number of weeks as 'w'. \textcolor{gray}{So, 16 times w equals 400.} In equation form, that's 16w = 400. \textcolor{gray}{To solve for 'w', I need to divide both sides of the equation by 16.} So, w equals 400 divided by 16. \textcolor{gray}{Let me calculate that.} \textcolor{gray}{400 divided by 16.} \textcolor{gray}{Hmm, 16 times 25 is 400 because 16 times 20 is 320, and 16 times 5 is 80, so 320 plus 80 is 400.} So, 16 times 25 is 400, which means w is 25. Wait, let me double-check that division to make sure I didn't make a mistake. 16 goes into 400 how many times? \textcolor{gray}{16 times 10 is 160, so 400 minus 160 is 240.} 16 times 15 is 240, so 10 plus 15 is 25. \textcolor{gray}{Yep, that's correct.} \textcolor{gray}{So, 16 times 25 is indeed 400.} \textcolor{gray}{Therefore, Chandler needs to work for 25 weeks to earn the remaining \$400 after using his birthday money.} \textcolor{gray}{So, in 25 weeks, he'll have enough to buy the mountain bike.} \textcolor{gray}{Just to recap, Chandler has \$100 from birthday gifts, needs \$500 total, so he needs an additional \$400.} He earns \$16 each week, so dividing 400 by 16 gives 25 weeks. That seems right. I don't think I made any mistakes here. \textcolor{gray}{The calculations are straightforward, and the division checks out.} So, I'm confident that the answer is 25 weeks. **Final Answer** Chandler will be able to buy the mountain bike in \textbackslash boxed\{25\} weeks.
  }
\end{qabox}

\begin{qabox}{Olmo-3-7B-Think, HARP}
    \textbf{Input:} Solve the following problem. You MUST put your final answer in \textbackslash boxed\{\}. \\
    Problem: Chandra pays an on-line service provider a fixed monthly fee plus an hourly charge for connect time. Her December bill was $\$12.48$, but in January her bill was $\$17.54$ because she used twice as much connect time as in December. What is the fixed monthly fee?
  \tcblower          %
  \textbf{Reasoning Chain:} 
    \texttt{
\textbf{Okay, so I need to figure out the fixed monthly fee that Chandra pays.}
\textbf{The problem says she has a fixed monthly fee plus an hourly charge for connect time.} 
\textbf{In December, her bill was \$12.48, and in January, it was \$17.54 because she used twice as much connect time as in December.} 
\textbf{Hmm, let me break this down.} 
\textbf{\textcolor{OliveGreen}{First, let me assign some variables to the unknowns.}} 
\textbf{Let's say the fixed monthly fee is F dollars.} 
\textbf{That's the part that's the same every month, right?} 
\textbf{Then there's the hourly charge.} 
\textbf{Let me call the hourly rate H dollars per hour.} 
\textbf{So, the total bill each month is the fixed fee plus the hourly rate multiplied by the number of hours she used that month.} 
\textbf{In December, her bill was \$12.48.} 
\textbf{Let me denote the number of hours she used in December as C hours.} 
\textbf{Then, the equation for December's bill would be:} 
\textbf{\boldmath{$F + H * C = 12.48$}} 
\textbf{\textcolor{orange}{Then in January, she used twice as much connect time, so that would be 2C hours.}} 
\textbf{\textcolor{orange}{And her January bill was \$17.54.}} 
\textbf{\textcolor{orange}{So the equation for January would be:}} 
\textbf{\boldmath{$F + H * (2C) = 17.54$}} 
\textbf{So now I have two equations:} 
\textbf{1) \boldmath{$F + H*C = 12.48$}} 
\textbf{\textcolor{OliveGreen}{2) \boldmath{$F + 2H*C = 17.54$}}} 
\textbf{I need to solve these two equations to find F, the fixed monthly fee.} 
\textbf{\textcolor{OliveGreen}{Since we have two equations and two variables (\boldmath{$F$} and \boldmath{$H*C$}), maybe I can subtract the first equation from the second to eliminate F.}} 
\textbf{Let me try that.} 
\textbf{Subtract equation 1 from equation 2:} 
\textbf{\textcolor{OliveGreen}{\boldmath{$(F + 2H*C) - (F + H*C) = 17.54 - 12.48$}}} 
\textbf{Simplify the left side:} 
\textbf{\textcolor{OliveGreen}{\boldmath{$F - F + 2H*C - H*C = 17.54 - 12.48$}}} 
\textbf{Which becomes:} 
\textbf{\textcolor{OliveGreen}{\boldmath{$H*C = 5.06$}}} 
\textbf{\textcolor{OliveGreen}{So, H*C is \$5.06.}} 
\textbf{\textcolor{OliveGreen}{Wait, but H is the hourly rate and C is the number of hours in December.}} 
\textbf{\textcolor{OliveGreen}{So \boldmath{$H*C$} is the variable cost for December's connect time.}} 
\textbf{\textcolor{OliveGreen}{Then, the fixed fee F can be found by plugging back into equation 1.}} 
\textbf{\textcolor{OliveGreen}{From equation 1: F + (H*C) = 12.48}} 
\textbf{We just found that H*C = 5.06, so substituting that in:} 
\textbf{\textcolor{OliveGreen}{\boldmath{$F + 5.06 = 12.48$}}}
\textcolor{gray}{Therefore, $F = 12.48 - 5.06$} 
Let me compute that. 12.48 minus 5.06. 
\textcolor{gray}{Hmm, $12.48 - 5$ is 7.48, then subtract another 0.06, so 7.42.} \textcolor{gray}{So F = \$7.42?}
Wait, let me check the subtraction again to be sure. 12.48 minus 5.06: 12.48 -5.06 -------- Let's do it step by step. 12.48 minus 5 is 7.48. \textcolor{gray}{Then subtract the remaining 0.06: 7.48 - 0.06 = 7.42.} Yeah, that's right. \textcolor{gray}{So F is \$7.42?} But let me verify with the second equation to make sure. \textcolor{gray}{If F is 7.42, then in January, the bill is F + 2*(H*C).} We know H*C is 5.06, so 2*(H*C) is 10.12. \textcolor{gray}{Then 7.42 + 10.12 should equal 17.54.} Calculating 7.42 + 10.12: 7 + 10 is 17, 0.42 + 0.12 is 0.54, so total is 17.54. Perfect, that matches the January bill. So that seems to check out. \textcolor{gray}{Therefore, the fixed monthly fee is \$7.42.} So I should put that in a box. Wait, but let me just make sure I didn't make any wrong assumptions. The problem says "a fixed monthly fee plus an hourly charge for connect time." So the total bill is fixed fee plus (hourly rate times hours used). That's how I set it up. So December: F + H*C = 12.48, January: F + H*(2C) =17.54. \textcolor{gray}{Then subtracting gives H*C = 5.06, so fixed fee is 12.48 -5.06=7.42.} Then verification works. Seems solid. Is there another way to approach this? Maybe solve for one variable in terms of the other? From the first equation, F = 12.48 - H*C. Then plug that into the second equation: 12.48 - H*C + 2H*C = 17.54 Simplify: 12.48 + H*C =17.54 Then H*C =17.54 -12.48 =5.06, same as before. \textcolor{gray}{So then F=12.48 -5.06=7.42.} Yep, same result. So that's consistent. Therefore, I think that's the right answer. \textcolor{gray}{The fixed monthly fee is \$7.42.} \textcolor{gray}{So the answer is 7.42, in a box.} **Final Answer** The fixed monthly fee is \textbackslash boxed\{7.42\}.
    }
\end{qabox}

\begin{qabox}{DeepSeek-R1-7B, MATH-500}
    \textbf{Input:} Solve the following problem. You MUST put your final answer in \textbackslash boxed\{\}. \\
    Problem: What is the integer value of $x$ in the arithmetic sequence $3^2, x, 3^4$?
  \tcblower          %
  \textbf{Reasoning Chain:} 
    \texttt{
\textbf{\textcolor{OliveGreen}{Okay, so I have this problem here: I need to find the integer value of x in the arithmetic sequence \boldmath{$3^2, x, 3^4$}.}} 
\textbf{\textcolor{orange}{Hmm, arithmetic sequence, right?}} 
\textbf{Let me think about what that means.} 
\textbf{\textcolor{OliveGreen}{An arithmetic sequence is a sequence of numbers where the difference between consecutive terms is constant.}} 
\textbf{\textcolor{OliveGreen}{So, if I have three terms in a row, the middle one is the average of the first and the third.}} 
\textbf{\textcolor{OliveGreen}{That makes sense because in an arithmetic sequence, each term is obtained by adding a common difference.}} 
\textbf{\textcolor{OliveGreen}{So, if I have the first term, then the second term is first term plus d, and the third term is second term plus d, which is first term plus 2d.}} 
\textbf{\textcolor{OliveGreen}{So, in this case, the first term is \boldmath{$3^2$}, which is 9}.}
\textbf{\textcolor{OliveGreen}{The third term is \boldmath{$3^4$}, which is 81.}} 
\textbf{\textcolor{OliveGreen}{So, the sequence is 9, x, 81.}} 
\textbf{\textcolor{OliveGreen}{Since it's an arithmetic sequence, the difference between the first and second term should be the same as the difference between the second and third term.}} 
\textbf{\textcolor{OliveGreen}{Let me write that down.}} 
\textbf{\textcolor{OliveGreen}{The difference between the second term and the first term is \boldmath{$x - 9$}.}} 
\textbf{\textcolor{OliveGreen}{The difference between the third term and the second term is \boldmath{$81 - x$}.}} 
\textbf{\textcolor{OliveGreen}{Since these differences are equal, I can set up the equation:}} 
\textbf{\textcolor{gray}{\boldmath{$x - 9 = 81 - x$}}} 
\textbf{\textcolor{OliveGreen}{Okay, so now I need to solve for x.}} 
\textbf{\textcolor{orange}{Let me do that step by step.}} 
\textbf{\textcolor{OliveGreen}{First, I can add x to both sides to get rid of the x on the right side.}} 
\textbf{\textcolor{orange}{So, adding x to both sides:}} 
\textbf{\textcolor{OliveGreen}{\boldmath{$x - 9 + x = 81 - x + x$}}} 
\textbf{\textcolor{OliveGreen}{Simplifying both sides:}} \textcolor{gray}{2x - 9 = 81}
Now, I can add 9 to both sides to isolate the term with $x$: $2x - 9 + 9 = 81 + 9$ 
\textcolor{gray}{Which simplifies to:  $2x = 90$}
Then, I can divide both sides by $2$ to solve for $x$: 
\textcolor{gray}{$2x / 2 = 90 / 2$} 
\textcolor{gray}{So, $x = 45$}
Wait, let me check if that makes sense.
\textcolor{gray}{If $x$ is $45$, then the sequence is $9, 45, 81$.} Let's see if the differences are the same. $45 - 9 = 36$, $81 - 45 = 36$ Yes, both differences are $36$, so that works.
\textcolor{gray}{So, $x$ is $45$.}
But hold on, the problem says to put the final answer in a box and as an integer.
\textcolor{gray}{$45$ is an integer, so that should be fine.}
Just to make sure I didn't make any mistakes, let me go through the steps again. First term: $3^2 = 9$ Third term: $3^4 = 81$ In an arithmetic sequence, the middle term is the average of the first and third terms. So, $x$ should be $(9 + 81)/2$. Let me calculate that: $(9 + 81) = 90$ 
\textcolor{gray}{$90$ divided by $2$ is $45$.}
Yep, that's the same answer I got before.
So, that confirms it.
I think that's solid.
I don't see any mistakes in my reasoning.
\textcolor{gray}{So, $x$ is definitely $45$.}
**Final Answer** The integer value of $x$ is \textbackslash boxed\{$45$\}.
    }
\end{qabox}

\end{document}